\documentclass{article}

\PassOptionsToPackage{numbers, compress}{natbib}

    \usepackage[preprint]{neurips_2025}

\usepackage[table]{xcolor}         
\usepackage[most]{tcolorbox}
\usepackage[utf8]{inputenc} 
\usepackage[T1]{fontenc}    
\usepackage{hyperref}       
\usepackage{url}            
\usepackage{booktabs}       
\usepackage{amsfonts}       
\usepackage{nicefrac}       
\usepackage{microtype}      
\usepackage{makecell}
\usepackage{multirow}
\usepackage{pifont}
\usepackage{subcaption}
\usepackage{graphicx}
\usepackage{textgreek}
\usepackage{wrapfig} 
\usepackage{pdfpages}
\usepackage{array}   
\usepackage{caption} 
\title{RSCC: A Large-Scale Remote Sensing Change Caption Dataset for Disaster Events}
\definecolor{tech_purple}{RGB}{168,119,200}
\definecolor{green}{RGB}{112,173,71} 
\definecolor{blue}{rgb}{0, 0, 1} 
\definecolor{orange}{rgb}{1, 0.27, 0} 
\definecolor{red}{rgb}{1, 0, 0} 

\makeatletter
\renewcommand\@fnsymbol[1]{\ensuremath{\ifcase#1\or *\or \dagger\or \ddagger\or \mathsection\or \mathparagraph\or \|\or **\or \dagger\dagger\or \ddagger\ddagger\else\@ctrerr\fi}}
\makeatother

\author{
  \parbox{\dimexpr\textwidth-2\tabcolsep\relax}{\centering
    {\bfseries Zhenyuan Chen\textsuperscript{1}\thanks{First author} \quad
    Chenxi Wang\textsuperscript{2} \quad
    Ningyu Zhang\textsuperscript{2} \quad
    Feng Zhang\textsuperscript{1,3,4}\thanks{Corresponding author}} \\
    {\normalfont\footnotesize \texttt{bili\_sakura@zju.edu.cn} \quad \texttt{sunnywcx@zju.edu.cn}} \\
    {\normalfont\footnotesize \texttt{zhangningyu@zju.edu.cn} \quad \texttt{zfcarnation@zju.edu.cn}} \\
    {\normalfont\footnotesize \textsuperscript{1}School of Earth Sciences, Zhejiang University, Hangzhou 310058, China \\
    \textsuperscript{2}School of Software Technology, Zhejiang University} \\
    {\normalfont\footnotesize \parbox{0.92\linewidth}{\centering
      \textsuperscript{3}Zhejiang Provincial Key Laboratory of Geographic Information Science, Hangzhou 310058, China\\
      \textsuperscript{4}Key Laboratory of Spatio-temporal Information and Intelligent Services (LSIIS), Ministry of Natural Resources of the People's Republic of China}}
    \\
  }
}

\begin{document}

\maketitle

\begin{abstract}
Remote sensing is critical for disaster monitoring, yet existing datasets lack temporal image pairs and detailed textual annotations. While single-snapshot imagery dominates current resources, it fails to capture dynamic disaster impacts over time. To address this gap, we introduce the Remote Sensing Change Caption (RSCC) dataset, a large-scale benchmark comprising 62,351 pre-/post-disaster image pairs (spanning earthquakes, floods, wildfires, and more) paired with rich, human-like change captions. By bridging the temporal and semantic divide in remote sensing data, RSCC enables robust training and evaluation of vision-language models for disaster-aware bi-temporal understanding. Our results highlight RSCC’s ability to facilitate detailed disaster-related analysis, paving the way for more accurate, interpretable, and scalable vision-language applications in remote sensing. Code and dataset are available at~\url{https://github.com/Bili-Sakura/RSCC}.
\end{abstract}

\begin{figure}[htbp]
\centering
\includegraphics[width=0.77\linewidth]{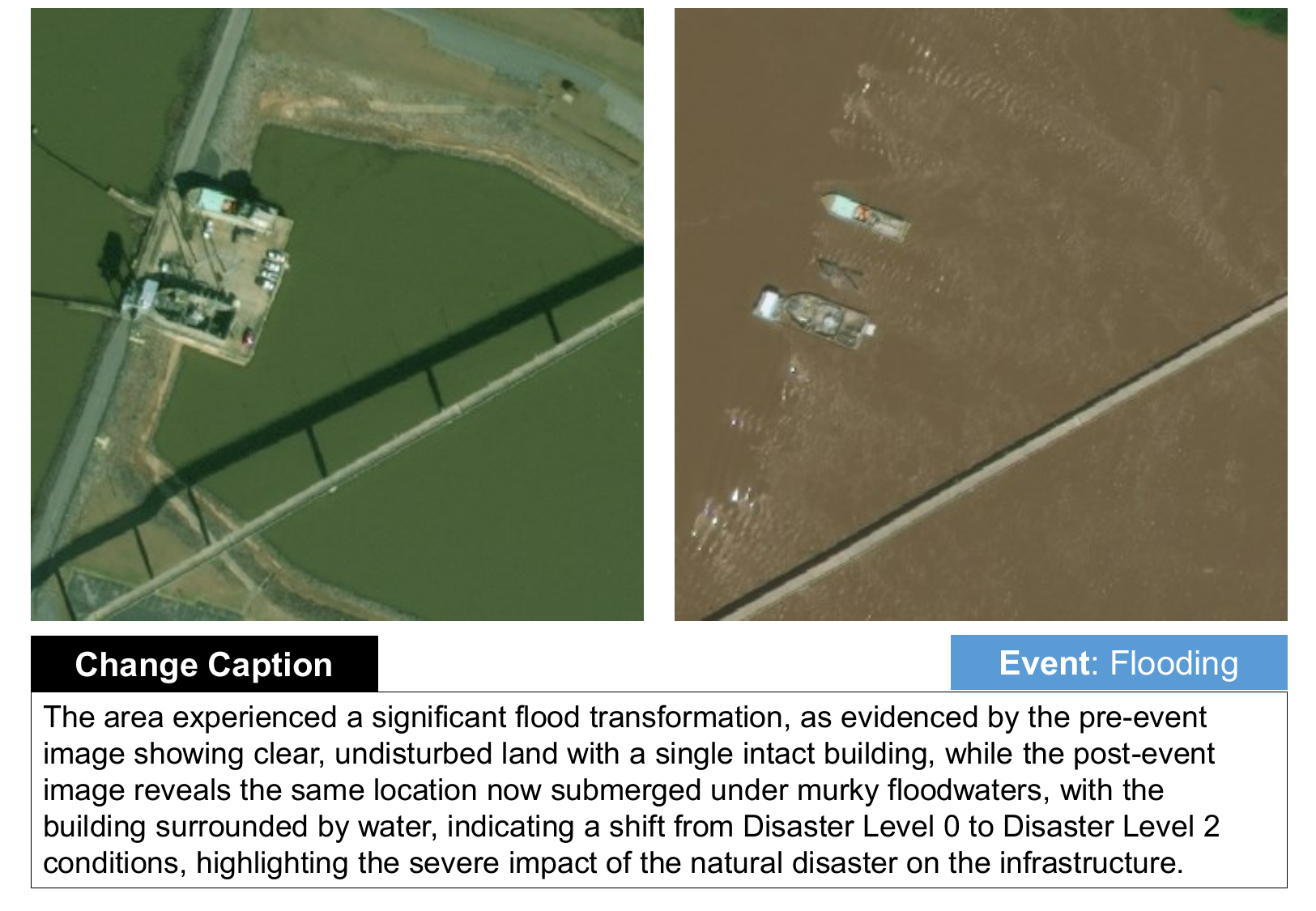}
\caption{A sample from RSCC dataset.}
\label{fig:data_overview}
\end{figure} 

\definecolor{forestgreen}{RGB}{34, 139, 34}

\section{Introduction}
\label{sec:intro}
Temporal remote sensing imagery is indispensable for monitoring dynamic Earth processes, particularly disaster events that demand rapid response and analysis. Temporal remote sensing data has proven indispensable in supporting disaster relief planning and response~\citep{rahnemoonfarRescueNetHighResolution2023,rahnemoonfarFloodNetHighResolution2021,guptaCreatingXBDDataset2019}. However, the inherently complex spatiotemporal relationships embedded within this data pose significant challenges for effective analysis and interpretation.

Advancements in the modeling of multimodal data have enabled generalist Multimodal Large Language Models (MLLMs)~\citep{anthropic2024claude,anthropicClaude3Model2024,openaiGPT4oSystemCard2024,openaiGPT4VisionSystemCard2023,teamGeminiFamilyHighly2024,teamGemini15Unlocking2024,grattafioriLlama3Herd2024,baiQwenVLVersatileVisionLanguage2023,wangQwen2VLEnhancingVisionLanguage2024a,llama4blog} that can perform a variety of natural image interpretation tasks specified flexibly through natural language. Specifically, MLLMs trained in a interleaved way have a deep visual-semantic understanding across images~\citep{chenExpandingPerformanceBoundaries2024a,liLLaVAOneVisionEasyVisual2024a,agrawalPixtral12B2024,wangQwen2VLEnhancingVisionLanguage2024,xueXGenMMBLIP3Family2024,liLLaVANeXTInterleaveTacklingMultiimage2024}. These models achieve great success in multi-image reasoning~\citep{mengMMIUMultimodalMultiimage2024,wangMuirBenchComprehensiveBenchmark2024,zhaoBenchmarkingMultiImageUnderstanding2024} and video understanding~\citep{liuBenchOpenEndedEventLevel2024,chandrasegaranHourVideo1HourVideoLanguage2024,caiTemporalBenchBenchmarkingFinegrained2024,heMMWorldMultidisciplineMultifaceted2024,fuVideoMMEFirstEverComprehensive2024}, while their capabilities in temporal remote sensing image understanding remain underexplored. 

Existing remote sensing image-text datasets often focus on single-snapshot imagery and lack the temporal details vital for understanding dynamic events, particularly in disaster-related scenarios, As shown in Table~\ref{tab:datasets}. Although there are multi-temporal datasets (e.g ., fMoW~\citep{christieFunctionalMapWorld2018}, SpaceNet~7~\citep{vanettenMultiTemporalUrbanDevelopment2021a}, S2Looking~\citep{shenS2LookingSatelliteSideLooking2021}, QFabric~\citep{vermaQFabricMultiTaskChange2021} and SpaceNet~8~\citep{hanschSpaceNet8Detection2022}), none of them provide rich textual descriptions of how scenes change over time. However, their potential for disaster-specific temporal analysis remains untapped due to the absence of high-quality bi-temporal datasets with detailed textual annotations. Existing remote sensing datasets either focus on generic land-use changes or provide short captions lacking disaster context. For instance, LEVIR-CC~\citep{liu_remote_2022} annotates urban development but omits disaster-specific details, while Dubai-CCD~\citep{hoxha_change_2022} offers brief descriptions without capturing nuanced damage levels or infrastructure transformations.

\begin{table}[!h]
\centering
\caption{Comparison with existing  remote sensing text-image datasets.}
\vspace{0.5em}
\renewcommand{\arraystretch}{1.25}
\begin{tabular}{lcrrcc}
\hline
\multirow{2}{*}{\textbf{Dataset}} & \multirow{2}{*}{\textbf{Year}} & \multirow{2}{*}{\textbf{\#Image (Pixels)}} & \multicolumn{2}{c}{\textbf{Caption}} & \multirow{2}{*}{\textbf{Temporal}} \\
\cline{4-5}
 & & & \textbf{\#Captions (Avg\_L)} & \textbf{Details} & \\
\hline
UCM-Captions~\citep{quDeepSemanticUnderstanding2016a}  & 2016 & 2,100 (1.0B) & 10,500 (12) & \textcolor{red}{\ding{55}} & \textcolor{red}{\ding{55}} \\
RSICD~\citep{luExploringModelsData2018a} & 2018 & 10,921 (0.5B) & 54,605 (12) & \textcolor{red}{\ding{55}} & \textcolor{red}{\ding{55}} \\
fMoW~\citep{christieFunctionalMapWorld2018} & 2018 & 1M (437.0B) & N/A & \textcolor{red}{\ding{55}} & \textcolor{forestgreen}{\checkmark} \\
SpaceNet~7~\citep{vanettenMultiTemporalUrbanDevelopment2021a} & 2021 & 2,389 (2.6B) & N/A & \textcolor{red}{\ding{55}} & \textcolor{forestgreen}{\checkmark} \\
S2Looking~\citep{shenS2LookingSatelliteSideLooking2021} & 2021 & 5,000 (5.0B) & N/A & \textcolor{red}{\ding{55}} & \textcolor{forestgreen}{\checkmark} \\
QFabric~\citep{vermaQFabricMultiTaskChange2021} & 2021 & 2,520 (245.1B) & N/A & \textcolor{red}{\ding{55}} & \textcolor{forestgreen}{\checkmark} \\
SpaceNet~8~\citep{hanschSpaceNet8Detection2022} & 2022 & 2,576 (3.0B) & N/A & \textcolor{red}{\ding{55}} & \textcolor{forestgreen}{\checkmark} \\
LEVIR-CC~\citep{liu_remote_2022} & 2022 & 20,154 (1.2B) & 50,385 (40) & \textcolor{forestgreen}{\checkmark} & \textcolor{forestgreen}{\checkmark} \\
Dubai-CCD~\citep{hoxha_change_2022} & 2022 & 1,000 (<0.1B) & 2,500 (35) & \textcolor{forestgreen}{\checkmark} & \textcolor{forestgreen}{\checkmark} \\
RSICap~\citep{huRSGPTRemoteSensing2023} & 2023 & 2,585 (0.6B) & 2,585 (60) & \textcolor{forestgreen}{\checkmark} & \textcolor{red}{\ding{55}} \\
RS5M~\citep{zhangRS5MGeoRSCLIPLargeScale2024} & 2024 & 5M (-) & 5M (49) & \textcolor{forestgreen}{\checkmark} & \textcolor{red}{\ding{55}} \\
VRSBench~\citep{li_vrsbench_2024} & 2024 & 29,614 (7.8B) & 29,614 (52) & \textcolor{forestgreen}{\checkmark} & \textcolor{red}{\ding{55}} \\
WHU-CDC~\citep{shiMultitaskNetworkTwo2024} & 2024 & 14,868 (1.9B) & 37,170 (-) & \textcolor{forestgreen}{\checkmark} & \textcolor{forestgreen}{\checkmark} \\
XLRS-Bench~\citep{wangXLRSBenchCouldYour2025} & 2025 & 934 (67.5B) & 934 (379) & \textcolor{forestgreen}{\checkmark} & \textcolor{red}{\ding{55}} \\

\hline
\textbf{RSCC (Ours)} & 2025 & 124,702 (32.7B) & 62,351 (72) & \textcolor{forestgreen}{\checkmark} & \textcolor{forestgreen}{\checkmark} \\
\hline
\end{tabular}
\addtolength{\abovecaptionskip}{0.15cm}
\label{tab:datasets}
\end{table}

To address these challenges, we introduce the Remote Sensing Change Caption (RSCC) dataset, the first large-scale dataset tailored for disaster-aware bi-temporal understanding . RSCC bridges critical gaps by:

\begin{enumerate}
    \item Large-Scale Event-Driven Dataset : 62,351 pre-/post-disaster image pairs sourced from 31 global events, spanning earthquakes, floods, wildfires, and more.
    \item A specialized model for remote sensing change captioning: To validate the robustness of our dataset, we train a MLLM specialized for remote sensing change captioning based on RSCC dataset. The benchmark result shows that RSCC dataset enhance the capabilities of general MLLMs on remote sensing temporal image understanding.
    \item Change Caption Benchmark : We develop a change caption benchmark based on our RSCC dataset and evaluate the performance of several state-of-the-art temporal MLLMs.
\end{enumerate}

The remainder of this paper is organized as follows. In Section~\ref{sec:pipeline}, we detail the construction process of RSCC, including data sources and caption generation pipeline. Section~\ref{sec:dataset} introduce the our specialzed remote sensing change captioning model trained on RSCC dataset.  In Section~\ref{sec:evaluation}, we benchmark existing temporal MLLMs' change captioning capabilities on RSCC and presents both qualitative and quantitative results.

\definecolor{headergray}{HTML}{333333} 

\section{Pipeline}
\label{sec:pipeline}
\begin{figure*}[!h]
\centering
\fbox{\includegraphics[width=0.98\linewidth]{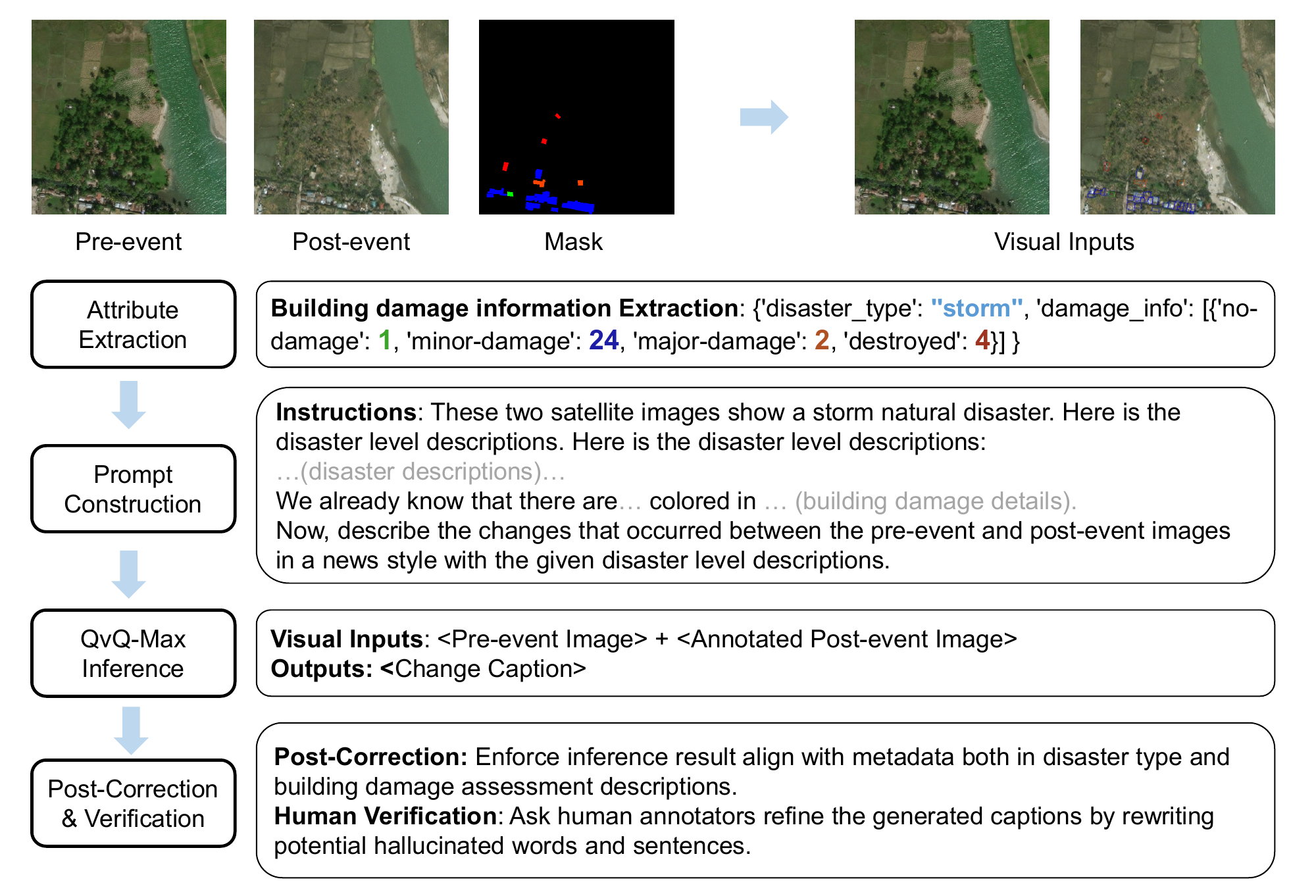}}
\caption{Illustration of RSCC dataset construction pipeline. We extract building damage information from labels and use carefully designed instructions to prompt QvQ-Max with reasoning capabilities and generate change captions from input images with building damage information.}
\label{fig:pipeline}
\end{figure*}

To construct our RSCC dataset, we employed a multimodal reasoning model - Qwen QvQ-Max~\citep{teamQVQMaxThinkEvidence2025} - along with existing human label to generate high fidelity captions. QvQ-Max is the latest proprietary MLLM that is capable of visual reasoning which shows superior capabilities in zero-shot remote sensing image change caption (see Appendix~\ref{appendix}). Unlike traditional MLLMs that prioritize recognition-based outputs, QvQ-Max leverages a structured reasoning process to infer spatial-temporal relationships~\cite{biVERIFYBenchmarkVisual2025}. The QvQ-Max captioning process takes about \$5/k image pairs. The overall dataset construction pipeline is shown in Figure~\ref{fig:pipeline}.

\subsection{Data Source}
In this study, we utilize xBD dataset~\citep{guptaCreatingXBDDataset2019} along with EBD dataset \citep{Wang2024EBD}, which are all obtained from MAXAR OpenData Program. The images are cropped without overlapping to 512×512 from xBD’s original 1024×1024, while EBD retains its 512×512 resolution. The overall RSCC datasets consists of 62,351 bi-temporal pre- and post-disaster image pairs (xBD:~44,136; EBD:~18,215) spanning from 31 events covering disaster types ranging from earthquake, flooding (hurricane), tsunami, storm (hurricane, tornado), volcano eruption and wildfire. Full events list is shown in Appendix~\ref{appendix}.

\subsection{Attribute Extraction}
The xBD dataset contains human annotations of building bounding boxes with damage assessment labels. The damage evaluation is based on the Joint Damage Scale~\citep{guptaCreatingXBDDataset2019}, which was developed with contributions from organizations such as NASA and the California Air National Guard. This scale is designed to assess building damage from satellite imagery across various disaster scenarios, providing detailed descriptions for different level ranging from no damage to destroyed.

\subsection{Prompt Construction}
\label{sec:prompt_construction}
We carefully design the following instructions to prompt QvQ-Max~\cite{teamQVQMaxThinkEvidence2025} to create detailed bi-temporal image change captions. We convert building damage labels into in-context auxiliary information. Shtedritski \textit{et al.}~\cite{shtedritskiWhatDoesCLIP2023} found that by applying marking-based visual prompt engineering, it is possible to unlock effective behaviors in vision-language models like CLIP~\cite{clip2021}, even without any training examples. This approach led to state-of-the-art results in zero-shot referring expression comprehension tasks. Inspired by this idea, we construct building damage masks as visual prompts for MLLMs. 

The prompt for QvQ-Max consists of visual inputs and textual inputs (instructions) (Figure~\ref{fig:pipeline}). The visual inputs are composed of original pre-event image and annotated post-event image where building bounding boxes are added onto the post-event image with different color that denote the damage level . The textual inputs are formatted as \textcolor{gray}{<task instructions> <disaster descriptions> <building damage details> }and \textcolor{gray}{<output format>} . The complete visual prompt template is shown in Appendix~\ref{appendix}.

\subsection{QvQ-Max Inference}

Given input prompts, we call QvQ-Max (qvq-max-2025-03-25) API from Alibaba Cloud \footnote{\url{https://bailian.console.aliyun.com/}} to automatically generate annotations. For xBD dataset change caption generation, we fix the prompt as one discussed in Section~\ref{sec:prompt_construction} which yield the optimal results in the empirical study. As EBD dataset does not contain human labeled annotation, we use naive prompt as "<pre\_image><post\_image>You will be provided with two satellite images of the same area before and after a \{disaster\_type\} natural disaster event. Describe the changes in a news style with a few sentences". We do not observe any issue of instruction mis-following or invalid output format for captions generated from both datasets.

\subsection{Post-Correction and Human Verification}
To ensure the reliability of captions generated by QvQ-Max, we implement a two-stage post-correction process. First, the Qwen2.5-Max~\citep{qwen_qwen25_2024} systematically enforces metadata alignment by correcting discrepancies in disaster type (e.g., resolving mismatches between "hurricane" and metadata-specified "flooding") and damage descriptions (e.g., revising "minor damage" to "destroyed" based on building annotations). This automated stage achieves disaster type consistency with metadata from 93.2\%  to 100.0\%.
Second, a subset of RSCC captions (10\%) is randomly selected and manually validated by three experts using a 0/1 binary rubric across four criteria: disaster type accuracy, damage detail completeness, factual consistency, and clarity. 100.0\% of sampled captions passed validation. Failed captions were reprocessed through the automated pipeline with refined rules, ensuring final dataset consistency. Full details of correction rules and evaluation protocols are provided in Appendix~\ref{appendix}.

\section{RSCC Dataset}
\label{sec:dataset}

\subsection{Overview}

\begin{figure}[!h]
    \centering
    \begin{subfigure}[b]{0.43\textwidth}
        \includegraphics[width=\textwidth]{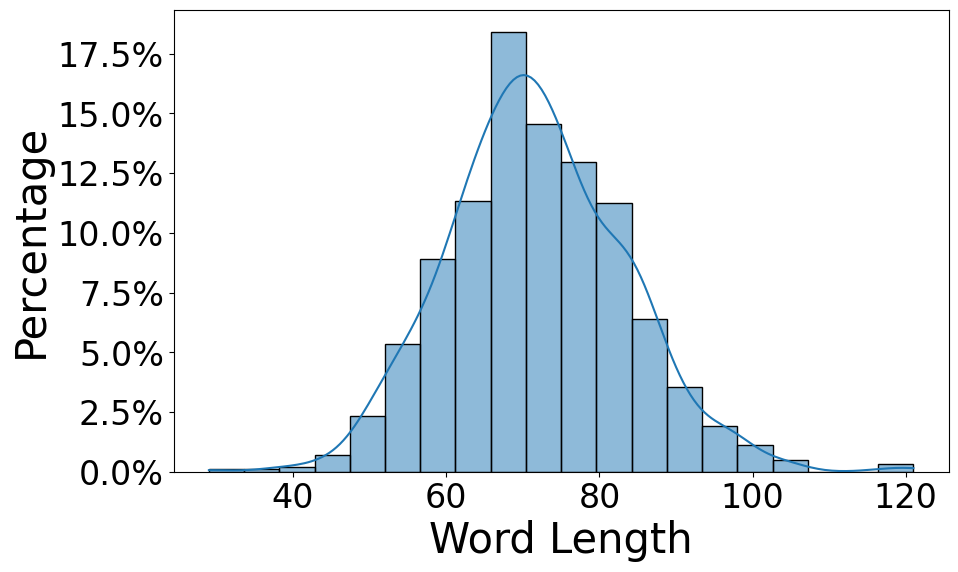}
        \caption{Number of words distribution.}
        \label{fig:subfig-a}
    \end{subfigure}
    \hfill
    \begin{subfigure}[b]{0.5\textwidth}
        \includegraphics[width=\textwidth]{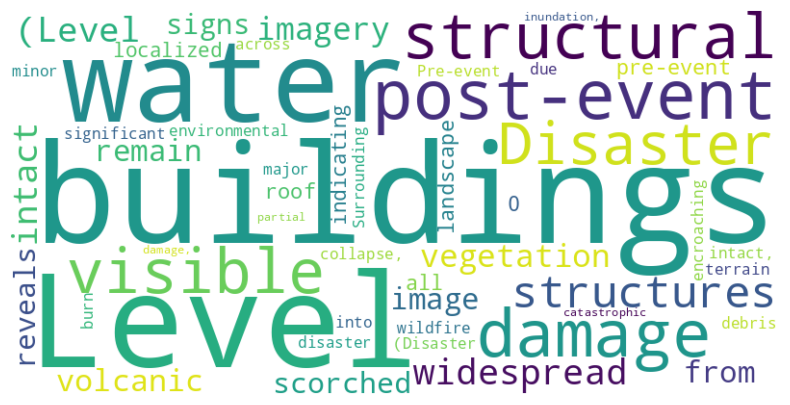}
        \caption{Wordcloud of captions.}
        \label{fig:subfig-b}
    \end{subfigure}
    \caption{Statistics of RSCC.}
    \label{fig:statistics}
\end{figure}

Our RSCC dataset comprises a total of 62,315 bi-temporal image pairs, each annotated with a detailed change caption. These image pairs capture a range of real-world disaster scenarios, reflecting a diverse set of geographical locations, disaster types, and severity levels. By offering rich textual descriptions of scene changes, RSCC aims to facilitate advanced temporal reasoning and caption generation tasks for large vision-language models. A summary of these caption statistics is detailed in Figure~\ref{fig:statistics}.

\subsection{RSCC for vision-language model training}
In order to facilitate the vision-language model training, we divide the RSCC dataset into two splits, where a \textit{train} split contains 61,363 image pairs from 31 distinct events across xBD as well as EBD and a test split contains 988 image pairs from 19 distinct events in xBD. We conduct full-parameter fine-tuning on Qwen2.5-VL 7B~\citep{baiQwen25VLTechnicalReport2025} using \textit{train} set for 2 epoch with a $batch\_size=1$ on a single node equipped with 2 NVIDIA H800 GPUs. We initialize the learning rate at 1e-6 and 1e-5 for LLM backbone and vision encoder respectively and employ a cosine learning rate decay schedule for optimization. For image inputs, we maintain the native resolution of RSCC as 512$\times$512 for maximum pixel inputs and minimum pixel inputs. The training procedure cost a total of 40 GPU clockwall hours.

\section{Benchmark Evaluation}
\label{sec:evaluation}

\subsection{Experiment Settings}

\paragraph{Baselines.}
For remote sensing change captioning, we benchmark moderate size open-sourced MLLMs (less than 13B parameters) that supports multi-image inputs, including LLaVA-NeXT-Interleave~\citep{liLLaVANeXTInterleaveTacklingMultiimage2024}, xGen-MM\footnote{\url{https://hugggingfae.co/Salesforce/xgen-mm-phi3-mini-instruct-interleave-r-v1.5}} (BLIP-3)~\citep{xueXGenMMBLIP3Family2024}, LLaVA-OneVision~\citep{liLLaVAOneVisionEasyVisual2024a}, Qwen2-VL~\citep{wangQwen2VLEnhancingVisionLanguage2024},Pixtral~\citep{agrawalPixtral12B2024} Phi-4-Multimodal~\citep{microsoftPhi4MiniTechnicalReport2025} Kimi-VL \citep{kimiteam2025kimivltechnicalreport} and InternVL 3~\citep{Internvl3}. 
We also add two specialized remote sensing change captioning models namely TEOChat~\citep{irvinTEOChatLargeVisionLanguage2024a} and CCExpert~\citep{wangCCExpertAdvancingMLLM2024}.
\paragraph{Evaluation Metrics.} For model evaluation, we compare the text similarity with n-gram overlap metrics including ROUGE~\citep{linROUGEPackageAutomatic2004} and METEOR~\citep{banerjeeMETEORAutomaticMetric2005}.  While the aforementioned measures are commonly reported in image captioning works, we find they are suboptimal to measure the semantic similarity across long texts. Therefore, we follow Kaggle LLM Prompt Recovery Competition \footnote{\url{https://www.kaggle.com/c/llm-prompt-recovery}} and introduce Sentence T5-XXL Embedding~\citep{niSentenceT5ScalableSentence2022} with Sharpened Cosine Similarity~\citep{Rohrer2022} (ST5-SCS) to get a well-established similarity measure. We set $q=0$ and $p=3$ for sharpened cosine similarity.

\paragraph{MLLMs Configurations.} For model generation, we use the default sampling strategy derived from configuration.   We use the same prompt style in section~\ref{sec:pipeline} for xBD dataset and omit building damage assessment information for EBD dataset. We compare the performance of change captions by three settings (i.e. zero-shot, textual prompt and visual prompt). We evaluate on RSCC \textit{test} set. More implement details are shown in Appendix~\ref{appendix}.

\subsection{Quantitative Results}

Our evaluation on the RSCC dataset reveals three primary insights into image captioning performance (Table \ref{tab:captioning_results}):  

1. \textbf{Model Scale vs. Performance }
The performance of vision-language models for remote sensing change captioning generally improves with increased parameter count, as seen in LLaVA-NeXT-Interleave (8B) achieving 46.99\% ST5-SCS and Qwen2-VL (7B) reaching 45.55\% ST5-SCS . However, Kimi-VL (3B) exceeds expectations with 51.35\% ST5-SCS , indicating that architectural optimizations or domain-specific tuning can mitigate limitations in model size. Larger proprietary models like InternVL 3 (8B) and Pixtral (12B) dominate metrics such as ROUGE (19.87\% ) and ST5-SCS (79.18\% ), though open-source models remain competitive baselines.

\begin{table}[!ht]
\caption{Detailed image caption performance on the subset of RSCC dataset (naive/zero-shot results). Avg\_L denotes the average word number of generated captions. \textbf{Boldface} indicates the best performance while \underline{underline} denotes the suboptimal performance. $^*$BLIP-3 and LLaVA-OneVision tend to repeat their answer endlessly, which cause large caption lengths.}
\vspace{0.5em}
\centering
\small
\setlength{\tabcolsep}{1.35pt}
\renewcommand{\arraystretch}{1.25}
\begin{tabular}{lrrrr}
\toprule
\multicolumn{1}{c}{\textbf{Model}} &
\multicolumn{2}{c}{\textbf{N-Gram}} &
\multicolumn{1}{c}{\textbf{Contextual Similarity}} &
\multirow{2}{*}{\textbf{Avg\_L}} \\
\cmidrule(lr){2-3} \cmidrule(lr){4-4}
\multicolumn{1}{c}{\textbf{(\#activate params)}} &
\textbf{ROUGE(\%)↑} &
\textbf{METEOR(\%)↑} &
\textbf{ST5-SCS(\%)↑} &
\\
\midrule
BLIP-3 (3B)~\citep{xueXGenMMBLIP3Family2024} & 4.53 & 10.85 & 44.05 & $^*$456 \\
Kimi-VL (3B)\citep{kimiteam2025kimivltechnicalreport} & 12.47 & 16.95 & 51.35 & 87 \\
Phi-4-Multimodal (4B) \citep{microsoftPhi4MiniTechnicalReport2025} & 4.09 & 1.45 & 34.55 & 7 \\
Qwen2-VL (7B)\citep{wangQwen2VLEnhancingVisionLanguage2024} & 11.02 & 9.95 & 45.55 & 42 \\

LLaVA-NeXT-Interleave (8B) \citep{liLLaVANeXTInterleaveTacklingMultiimage2024} & 12.51 & 13.29 & 46.99 & 57 \\
LLaVA-OneVision (8B)\citep{liLLaVAOneVisionEasyVisual2024a} & 8.40 & 10.97 & 46.15 & $^*$221 \\
InternVL 3 (8B) \citep{Internvl3} & \underline{12.76} & 15.77 & 51.84 & 64 \\
Pixtral (12B)~\citep{agrawalPixtral12B2024} & 12.34 & \underline{15.94} & 49.36 & 70 \\

\hline
\textcolor{gray}{CCExpert (7B)} \citep{wangCCExpertAdvancingMLLM2024} & 7.61 & 4.32 & 40.81 & 12 \\
\textcolor{gray}{TEOChat (7B)}\citep{irvinTEOChatLargeVisionLanguage2024a} & 7.86 & 5.77 & \underline{52.64} & 15 \\
\textbf{Ours (7B)} & \textbf{14.99} & \textbf{16.05} & \textbf{58.52} & 44 \\
\bottomrule
\end{tabular}
\label{tab:captioning_results}
\end{table}

2. \textbf{Specialized Model}
Specialized models fine-tuned on remote sensing data, including CCExpert (7B) , TEOChat (7B) , and Ours (7B) , exhibit mixed outcomes. Ours (7B) achieves 58.52\% ST5-SCS through targeted training on RSCC, outperforming general models like Qwen2-VL (7B) . In contrast, CCExpert and TEOChat underperform in completeness and accuracy despite their domain focus, highlighting challenges in handling complex spatiotemporal reasoning. Proprietary models like Pixtral (12B) and InternVL 3 (8B) set performance benchmarks, while general models like BLIP-3 (3B) struggle with excessive output length (Avg\_L=456 ) and low ROUGE scores (4.53\% ).


3. \textbf{Repetition Issue} 
     BLIP-3 and LLaVA-OneVision are prone to generative repetitive outputs.  It is assumed that these models fail in dealing with remote sensing images or following complex instructions. This degeneration problem may be alleviated by switching decoding methods (e.g., Contrastive Decoding~\cite{su_contrastive_2022}) as well as adapting generation configurations~\cite{welleck_neural_2019}.

\subsection{Human Preference Study}

\begin{figure*}[!h]
\centering
\includegraphics[width=1.0\linewidth]{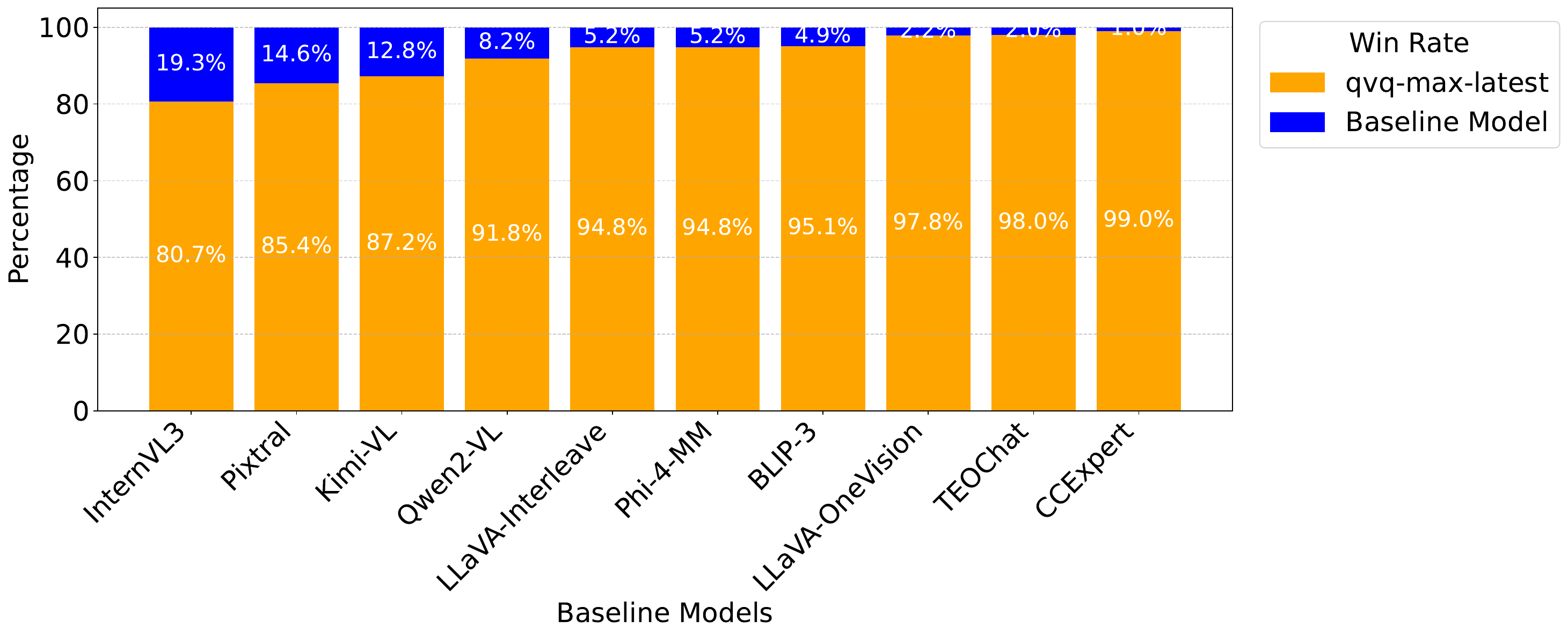}
\caption{Win-rate from QvQ-Max (ground truth) to all baseline models on RSCC subset.}
\label{fig:win_rate}
\end{figure*} 

While the language metrics can be biased, we ask experts to vote the best caption from two anonymous model output given the bi-temporal image pairs along with human labeled building damage masks from xBD dataset~\cite{guptaCreatingXBDDataset2019}. Results (Figure \ref{fig:win_rate}) reveal QvQ-Max (ground truth change captions) consistently outperformed all baselines, achieving win rates ranging from 80.7\% (against InternVL3) to 99.0\% (against CCExpert). While strong baselines like InternVL3 (\textbf{19.3\%} wins) and mid-tier models (e.g., Pixtral [\textbf{14.6\%}], Kimi-VL [\textbf{12.8\%}]) showed moderate performance, our captions demonstrated superior accuracy in capturing fine-grained environmental changes critical for disaster response. Weak-performing multimodal baselines (LLaVA-Interleave [\textbf{5.2\%}], Phi-4-MM [\textbf{4.9\%}]) highlighted limitations in handling complex spatiotemporal reasoning, suggesting QvQ-Max's quantization-aware training and dynamic context adaptation mechanisms enhance generalization. These findings validate QvQ-Max as a state-of-the-art solution for vision-language tasks in remote sensing.

\subsection{Inference-Time Augmentation}


\subsubsection{Employ Building Damage Info}

\begin{figure*}[!h]
\centering
\includegraphics[trim=0 300 0 10, clip,width=1.0\linewidth]{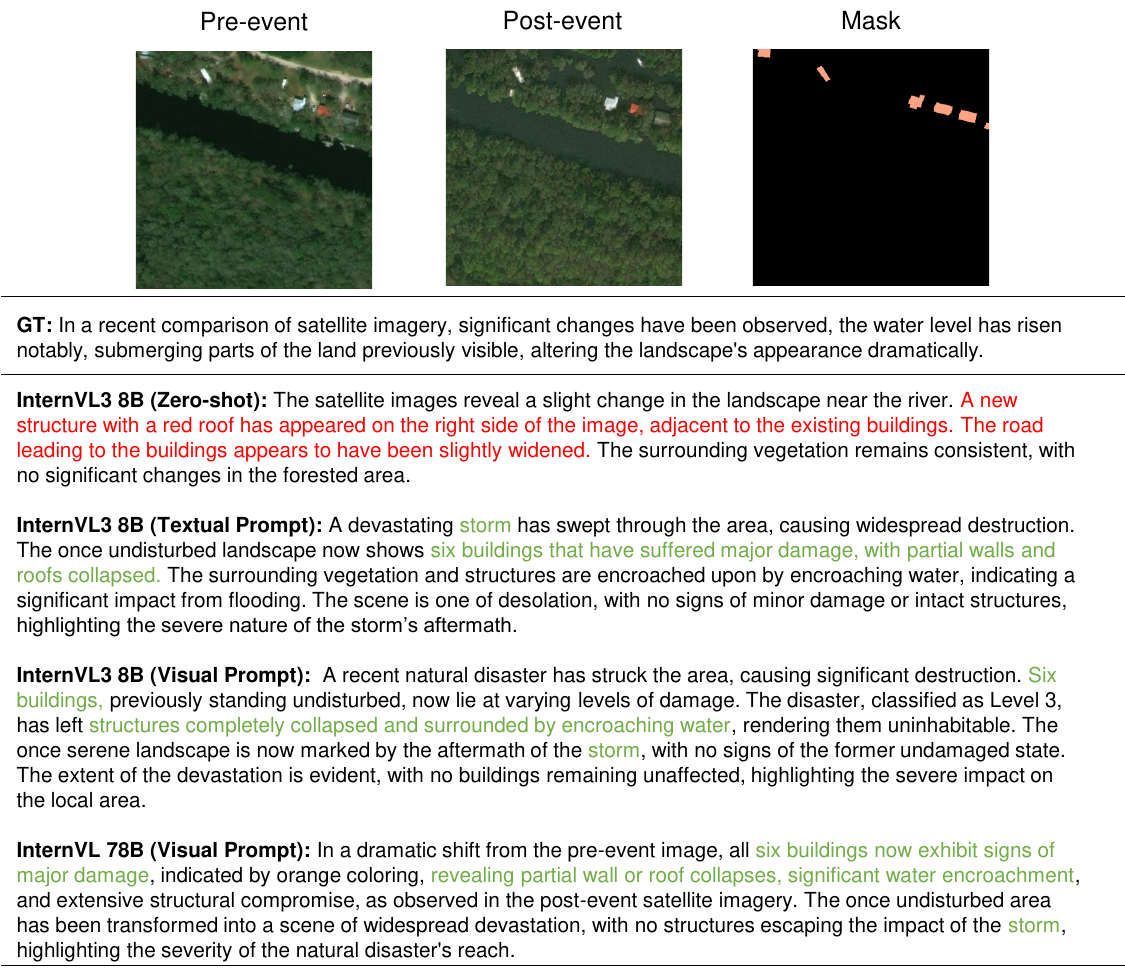}
\caption{Prompt augmentation results on RSCC (xBD: HURRICANE-FLORENCE). Critical descriptions are colored in \textcolor{green}{green} while incorrect and hallucinated sentences/words are \textcolor{red}{red}.}
\label{fig:augmentation}
\end{figure*}

Change caption result quality boost via augmentation with building damage info (Figure~\ref{fig:augmentation}). It is witnessed that \textbf{auxiliary building damage info augmentation greatly improve the quality of change captions. We also find performance gets saturated equipped with auxiliary info regardless model size} (see quantitative results in Table~\ref{tab:captioning_results_all} in Appendix~\ref{appendix}). 

\subsubsection{Scaling Correction Decoding}

\begin{figure*}[!h]
\centering
\begin{subfigure}[b]{0.485\linewidth} 
    \centering
    \fbox{\includegraphics[width=0.96\linewidth]{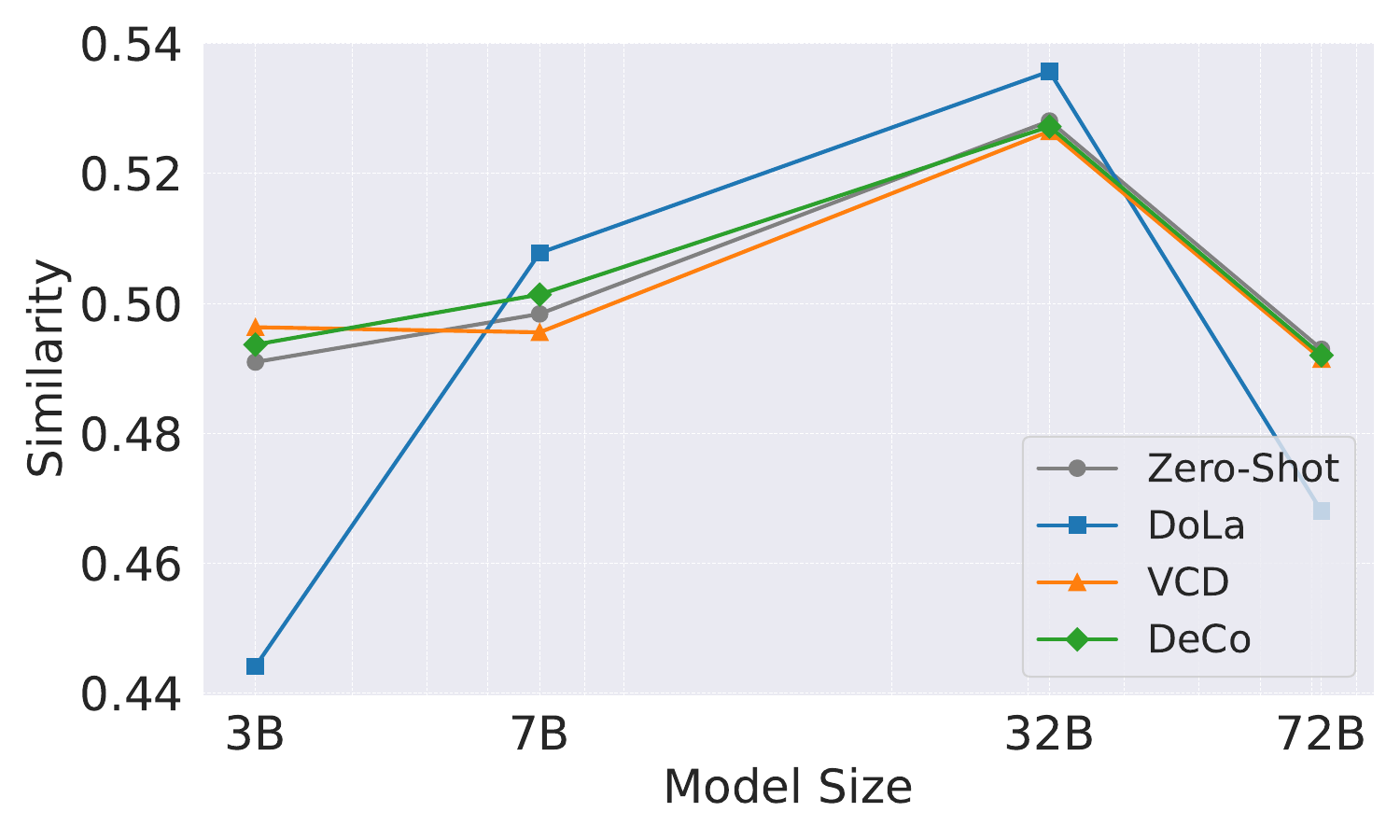}}
    \caption{Scaling correction decoding for Qwen2.5-VL}
    \label{fig:cd_scaling_qwenvl}
\end{subfigure}
\hfill 
\begin{subfigure}[b]{0.485\linewidth} 
    \centering
    \fbox{\includegraphics[width=0.96\linewidth]{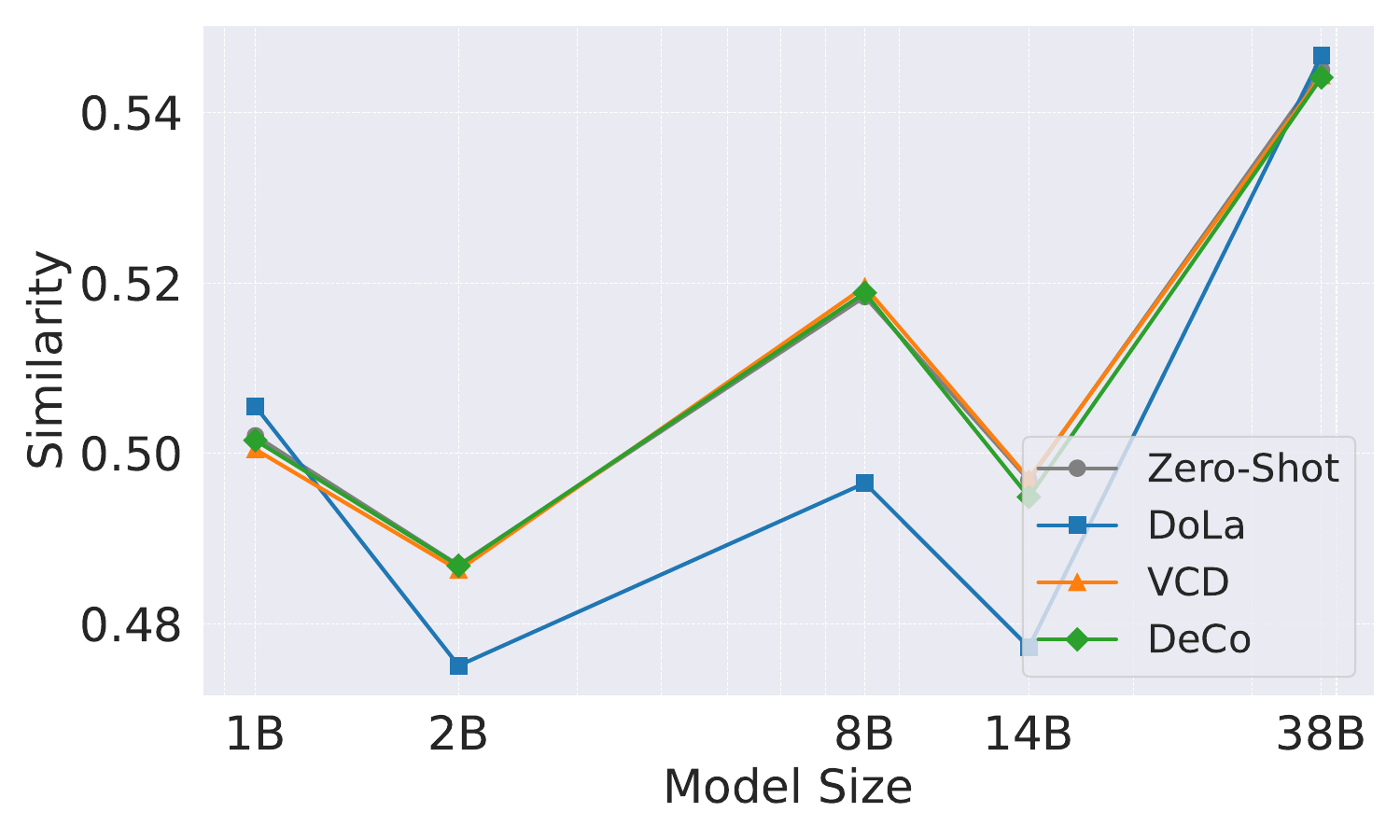}}
    \caption{Scaling correction decoding for InternVL3}
    \label{fig:cd_scaling_internvl}
\end{subfigure}
\caption{Comparison of scaling correction decoding}
\label{fig:cd_scaling}
\end{figure*}

To investigate the effectiveness of scaling correction decoding strategies (e.g., VCD~\cite{lengMitigatingObjectHallucinations2024}, DoLa~\cite{chuangDoLaDecodingContrasting2023} and DeCo~\cite{wangMLLMCanSee2024}) in mitigating hallucinations during remote sensing change captioning, we evaluated their impact across varying model sizes for Qwen2.5-VL and InternVL3 (Figure~\ref{fig:cd_scaling}). These strategies aim to align model outputs with input scale or context, reducing inconsistencies in multimodal reasoning.

\begin{figure*}[!h]
\centering
\includegraphics[trim=0 260 0 10, clip,width=1.0\linewidth]{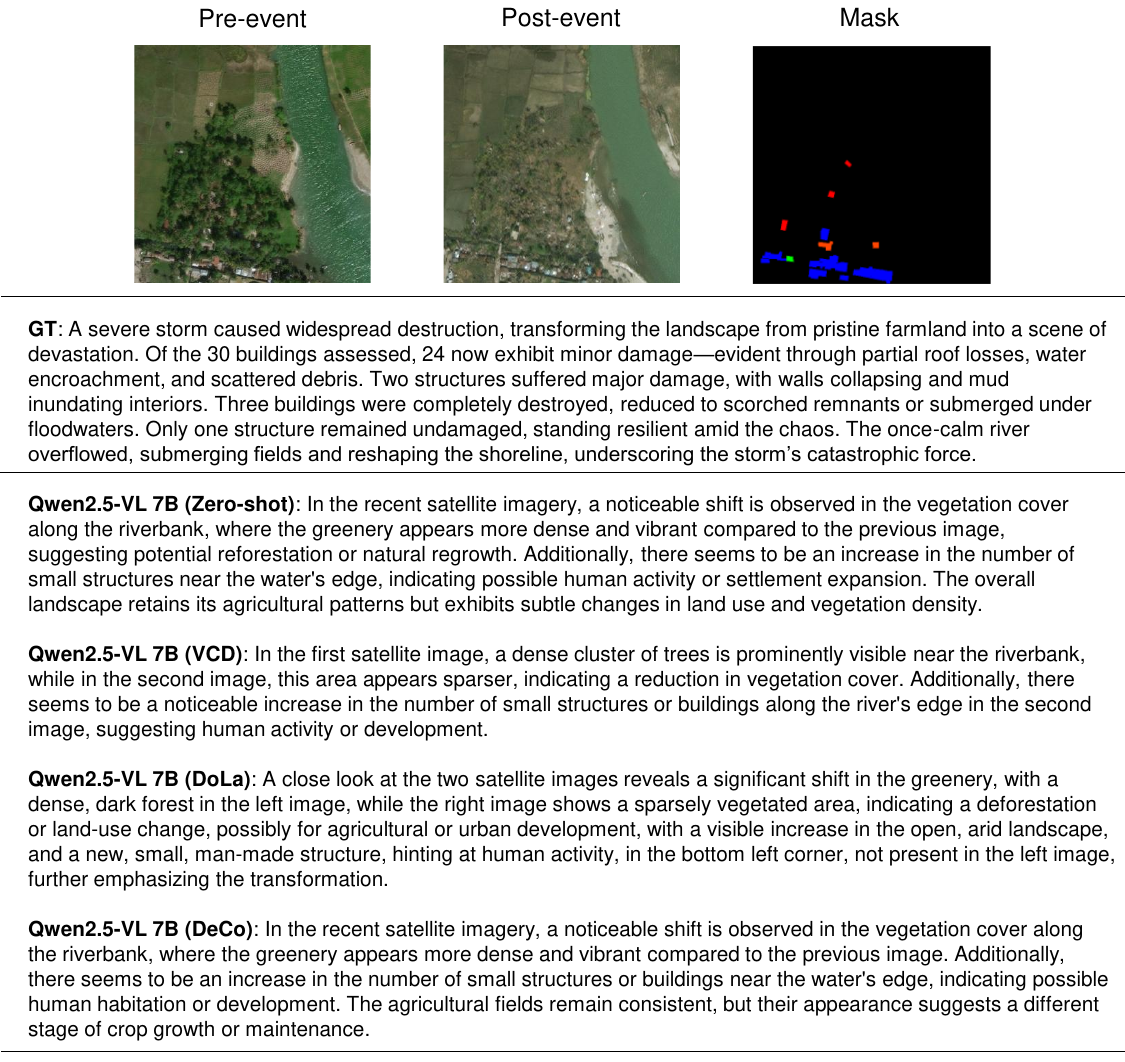}
\caption{Correction decoding results on RSCC (xBD: HURRICANE-MATTHEW).}
\label{fig:scaling_demo}
\end{figure*} 

For Qwen2.5-VL, Zero-Shot decoding achieves the highest similarity scores at smaller model sizes (3B–7B), while DeCo gradually closes the gap at larger scales (32B–72B). Notably, DoLa and VCD underperform across all sizes, suggesting limited utility for complex spatiotemporal reasoning tasks. In contrast, InternVL3 shows Zero-Shot decoding as the most consistent strategy, outperforming alternatives except at 14B, where DeCo marginally surpasses it. However, even at 38B parameters, scaling correction methods fail to achieve substantial gains over baseline performance. \textbf{We found no obvious boost using training-free correction decoding strategies on remote sensing change captioning task where we contribute to the abilities of complex visual reasoning instead of naive object level detection} (see Figure~\ref{fig:scaling_demo} for a case study).

\section{Related Work}
\label{sec:related}

\subsection{Remote Sensing Change Captioning Models}

The remote sensing image change captioning (RSICC) task aims to generate detailed and accurate natural language to describe geospatial feature changes for remote sensing images captured at different times~\citep{changChangesCaptionsAttentive2023,hoxhaChangeCaptioningNew2022}. Liu et al. \citep{liu_remote_2022} introduced RSICCformer, a Transformer-based approach that incorporates multiple cross-encoding modules to leverage differential features, focusing attention on the changing regions within each image. Similarly, Chg2Cap \cite{chang_changes_2023} presents a Transformer-based caption generation model that translates the relationship between image embeddings and word embeddings into descriptive text.
Recent researches tend to use pre-trained LLM for language generation. 
GeoLLaVA~\citep{elgendyGeoLLaVAEfficientFineTuned2024} use fine-tune pretrained temporal MLLMs (i.e. Video-LLaVA~\cite{linVideoLLaVALearningUnited2024} and LLaVA-NeXT-Video~\cite{liLLaVANeXTInterleaveTacklingMultiimage2024}) for detecting temporal changes in geographical landscapes. CCExpert~\cite{wangCCExpertAdvancingMLLM2024}, which is developed based on LLaVA-OneVision~\citep{liLLaVAOneVisionEasyVisual2024a}, introduces a difference-focused integration component. This module is designed to identify multi-scale variations between bi-temporal images and merge them into the initial image context. TEOChat~\cite{irvinTEOChatLargeVisionLanguage2024a} applys a shared vision-encoder to agument the temporal understanding capability of LLaVA-1.5~\cite{liuImprovedBaselinesVisual2024a}.
Despite the common architecture that intergrates a pre-trained LLM backbone and a vision encoder, Diffusion-RSCC~\citep{yu_diffusion-rscc_2024} utilize a probabilistic diffusion model for RSICC that focus on pixel-level differences under long time span.

\subsection{Remote Sensing Change Caption Dataset} 
Datasets that combine temporal and vision-language elements play a crucial role in training models to comprehend and merge temporal dynamics with linguistic information~\cite{Liu_Zhang_Chen_Wang_Zou_Shi_2024}.
Unlike the VQA dataset~\cite{irvinTEOChatLargeVisionLanguage2024a,elgendyGeoLLaVAEfficientFineTuned2024} that can be easily formatted through mask labels, the common practice of remote sensing change caption dataset is to further annotate existing change detection dataset with 5 sentences each image pairs, such as Dubai-CCD~\citep{hoxha_change_2022}, LEVIR-CC~\citep{liu_decoupling_2023} and  WHU-CDC~\citep{shiMultitaskNetworkTwo2024}. Given the generalization capability of commercial MLLMs, Wang \textit{et al.} ~\cite{wangCCExpertAdvancingMLLM2024} leveraged GPT-4o~\cite{openaiGPT4oSystemCard2024}, using the explicit information provided by the change masks to generate detailed change descriptions.

\section{Limitations and Discussions}
\label{sec:limitations}
Due to the lack of proficient labels and complexity of image pairs themselves, the generated captions may contains vague descriptions which is even hard for experts to clarify. Besides, we only employ text similarity metrics because existing image-to-text captioning metrics (e.g., FLEUR~\citep{lee_fleur_2024}, SPARC~\citep{jung_visual_2025} and G-VEval~\citep{tong_g-veval_2024}) only focus on single image which fail in multi-image scenario. We leave these parts for future work.

Also, our preliminary study have tested baseline VLMs on change detection and multi-label classification upon RSCC, where it show that naively employing VLMs such tasks yields much inferior results compared to specialized models~\citep{zheng2024segmentanychange, shiMultitaskNetworkTwo2024, ding2024scanformer}. Moreover, a recent paper \citep{fu2025hidden} denotes that VLMs' performance would degrade compared to their visual encoders only. Thus, we generally recommend using specialized models for such visual-centric tasks, and we hope the community will develop strong VLMs that are able to naively deal with these tasks.

\section{Conclusions}
\label{sec:conclusions}

In this work, we introduced RSCC, a large-scale event-driven remote sensing change caption dataset for disaster-awareness bi-temporal remote sensing image understanding. By leveraging visual reasoning model QvQ-Max, 62,351 pairs of pre-event and post-event images are annotated with detailed change caption. Furthermore, We established a comprehensive benchmark to facilitate the evaluation and development of large vision-language models in remote sensing change captioning. Our work focuses on promoting the training and evaluation of vision-language models for tasks related to understanding temporal remote sensing images.

\section{Acknowledgements}
We gratefully acknowledge the valuable suggestions and preliminary work contributed by student members Tesi Lin and Zeyu Zhang. We also thank Alibaba Cloud (Aliyun) for providing access to the Qwen API, which was instrumental in this project. 

This work was supported by the National Natural Science Foundation of China under Grants 42394060 and 42394062 as well as the National Key Research and Development Program of China under Grant 2019YFE0127400.

{
    \small
    \bibliographystyle{ieeenat_fullname}
    \bibliography{bib}
}
\newpage
\appendix
\newpage
\section{Appendix}
\label{appendix}

\begin{figure*}[!ht]
\centering
\fbox{\includegraphics[trim=0 60 0 0, clip, width=0.94\linewidth]{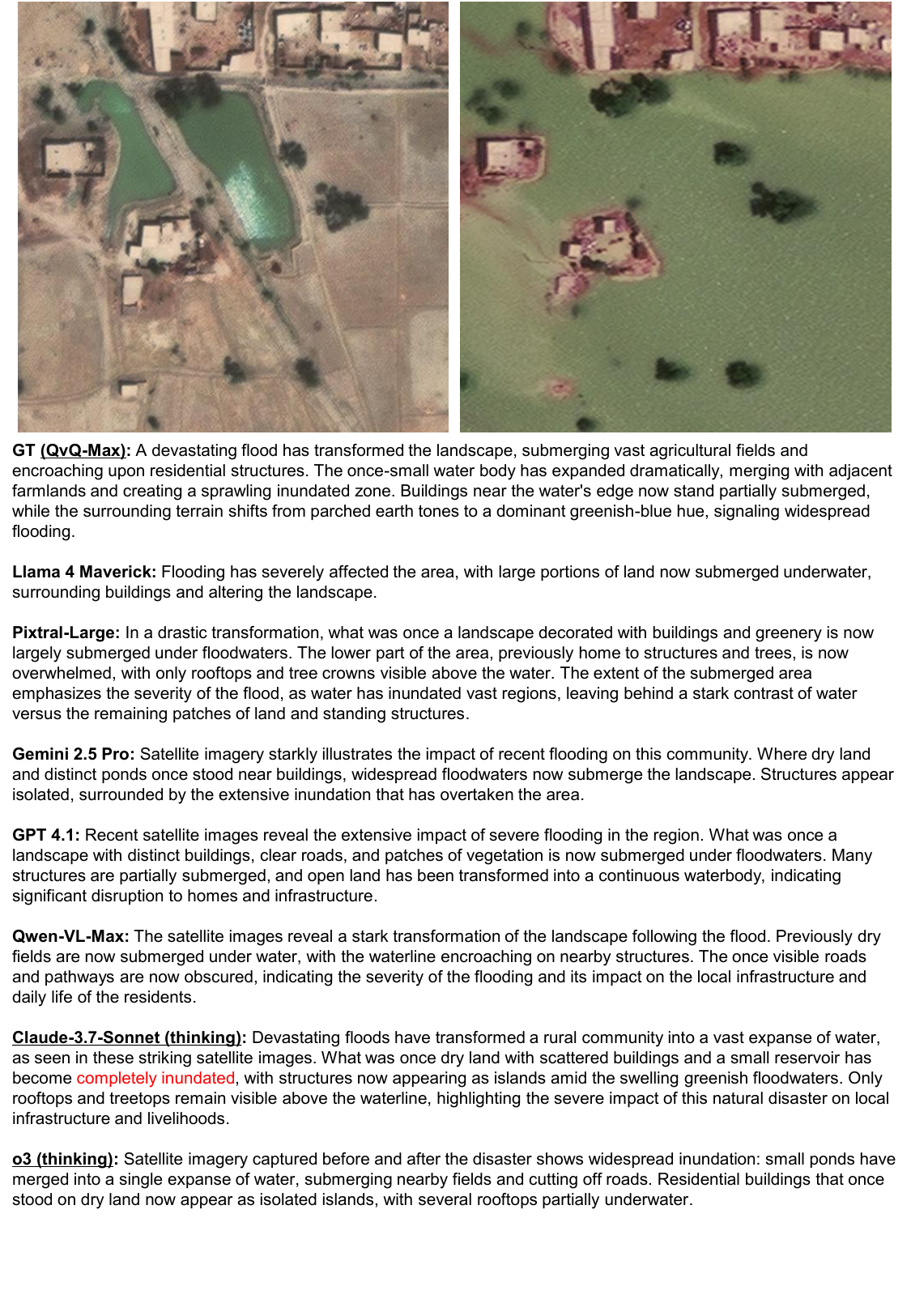}}
\caption{Comparisons of change captions of a pre-event image (Left) and a post-event image (Right) on RSCC (EBD: PAKISTAN-FLOODING) with Large Models . Words/sentences colored in \textcolor{green}{green}, \textcolor{red}{red} and \textcolor{tech_purple}{purple} denote to critical descriptions, incorrect descriptions and vague/undetermined descriptions respectively. Models with reasoning capabilities are \underline{underline}.}
\label{fig:vis_large_1}
\end{figure*} 

\newpage

\begin{figure*}[!ht]
\centering
\fbox{\includegraphics[trim=0 40 0 0, clip, width=0.98\linewidth]{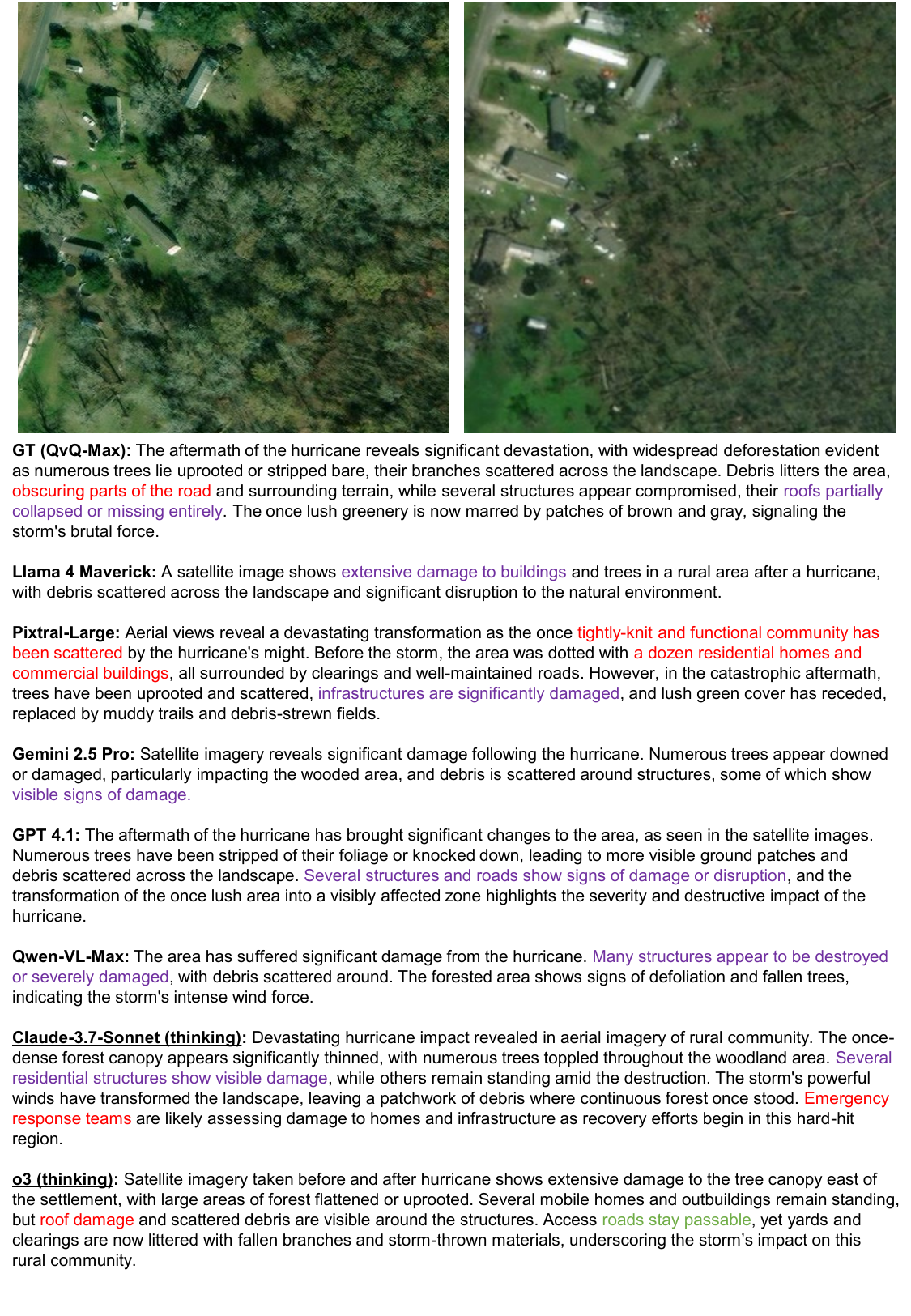}}
\caption{Comparisons of change captions of a pre-event image (Left) and a post-event image (Right) on RSCC (EBD: HURRICANE-IDA) with Large Models. Words/sentences colored in \textcolor{green}{green}, \textcolor{red}{red} and \textcolor{tech_purple}{purple} denote to critical descriptions, incorrect descriptions and vague/undetermined descriptions respectively. Models with reasoning capabilities are \underline{underline}.}
\label{fig:vis_large_2}
\end{figure*} 

\newpage

\begin{figure*}[!ht]
\centering
\fbox{\includegraphics[trim=0 0 0 0, clip, width=0.98\linewidth]{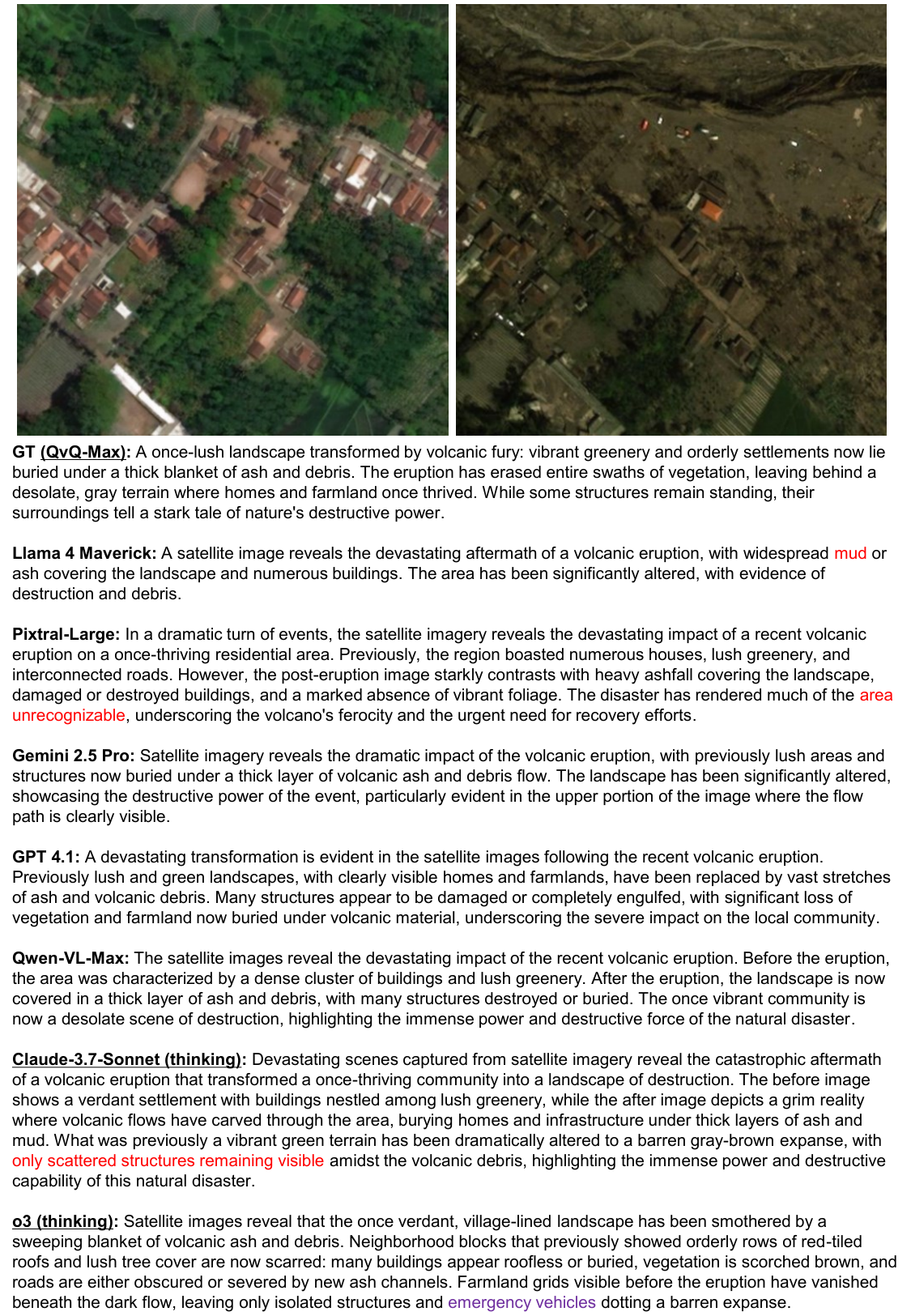}}
\caption{Comparisons of change captions of a pre-event image (Left) and a post-event image (Right) on RSCC (EBD: MOUNT-SEMERU-ERUPTION) with Large Models. Words/sentences colored in \textcolor{green}{green}, \textcolor{red}{red} and \textcolor{tech_purple}{purple} denote to critical descriptions, incorrect descriptions and vague/undetermined description respectively. Models with reasoning capabilities are \underline{underline}.}
\label{fig:vis_large_3}
\end{figure*} 

\newpage

\begin{figure*}[!ht]
\centering
\fbox{\includegraphics[width=0.98\linewidth]{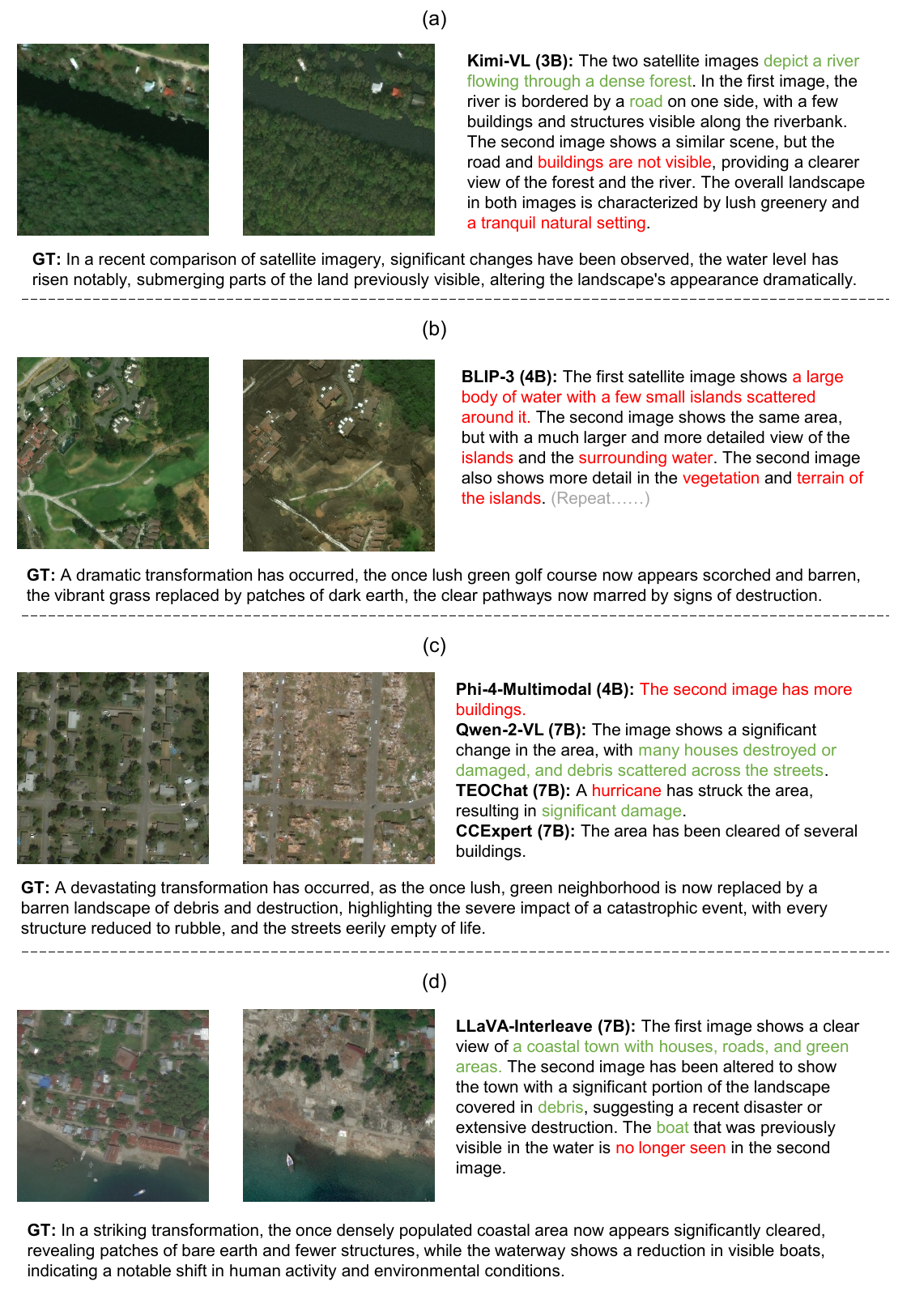}}
\caption{Visualization of qualitative results. Critical descriptions are colored in \textcolor{green}{green} while incorrect and hallucinated sentences/words are \textcolor{red}{red}. More results are shown in Figure~\ref{fig:more_examples}.}
\label{fig:comparison_examples}
\end{figure*} 

\newpage

\begin{figure*}[!ht]
\centering
\fbox{\includegraphics[trim=0 80 0 55, clip, width=0.98\linewidth]{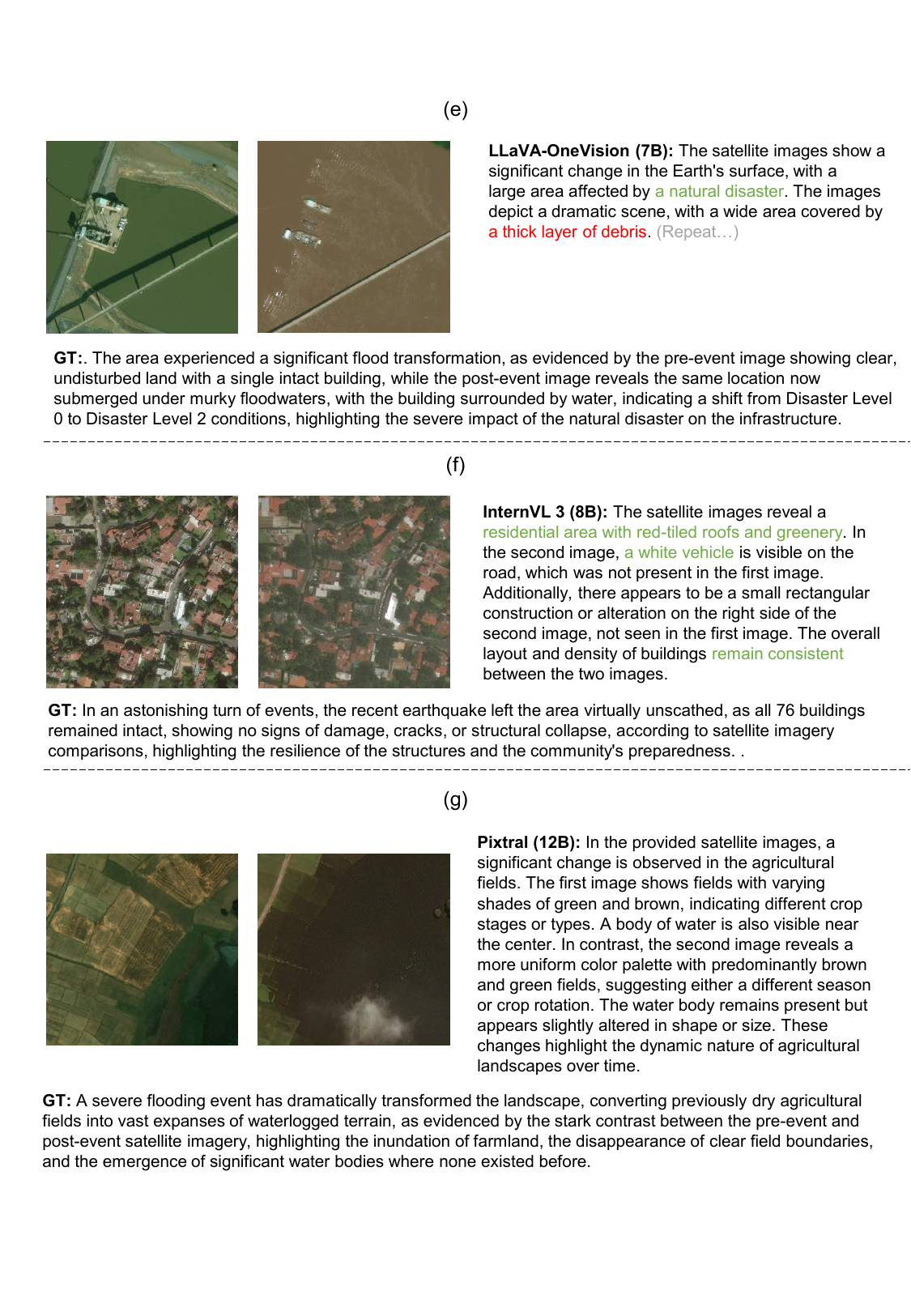}}
\caption{More examples of RSCC. Critical descriptions are colored in \textcolor{green}{green} while incorrect and hallucinated sentences/words are \textcolor{red}{red}.}
\label{fig:more_examples}
\end{figure*} 

\subsection{RSCC Captioning Details}
The experiments are implemented using the PyTorch framework and evaluated on an NVIDIA H800 GPUs (80GB). It takes about 1.1-8.3 seconds for captioning per image pair for all models with model size no more than 12B on a single H800 GPU.

We compare the performance of large-size MLLMs with zero-shot template (\ref{template}) including open-source models such as Pixtral Large~\citep{PixtralLargeMistral2024} and LLaMA-4 Maverick~\citep{llama4blog}. We also conduct case study on proprietary models including GPT-4.1 (2025-04-14)~\citep{openaiIntroducingGPT41API}, Gemeni-2.5-Pro (2025-03-25)~\citep{GeminiProGoogle}, and Qwen-VL Max (2025-01-25)~\citep{qwenteamIntroducingQwenVL}, along with reasoning model such as  Claude-3.7-Sonnet-Thinking (2025-02-25)~\citep{claude37} and o3 (2025-04-03) ~\citep{IntroducingOpenAIO3}. For results generation, We use default configurations of the above models. Figure~\ref{fig:vis_large_1},~\ref{fig:vis_large_2}~and~\ref{fig:vis_large_3} show qualitative results of empirical study. We found proprietary models outperform open-sourced models in completeness and accuracy. Visual reasoning notably improve the quality of caption in completeness but it also introduce vague information even hallucinations. 
As remote sensing change captioning requires world knowledge and complex reasoning, the latest state-of-the-art MLLMs seem to be insufficient.

\subsection{More Results}

Figure~\ref{fig:comparison_examples} presents a qualitative comparison of vision-language models across diverse remote sensing scenarios, highlighting their ability to detect and describe change. 

In Scenario (a), ground truth accurately identifies flooding as the disaster, highlights submergence of land, and links changes to water level rise, while Kimi-VL omits disaster causation and misrepresents structural disappearance as improved visibility. 

In Scenario (b), ground truth accurately identifies the disaster type (fire/heat damage) and captures key changes: scorched vegetation, dark earth replacing greenery, and damaged pathways. Its description aligns with typical wildfire impacts (burnt surfaces, structural debris) while BLIP-3 incorrectly references a "body of water" and "islands," which are absent in the images, failing basic accuracy and relevance. 

In Scenario (c), ground truth provides the most accurate, complete, and factually consistent description. It captures the catastrophic scale of destruction ("every structure reduced to rubble," "barren landscape"), explicitly mentions debris and empty streets, and aligns with typical patterns of severe wind-driven disasters (e.g., hurricanes or tornadoes). While it does not specify the disaster type, its focus on observable damage patterns (total structural collapse, vegetation loss) adheres strictly to visual evidence. Other captions either misinterpret the scene (Phi-4-MM, CCExpert), lack detail (TEOChat), or omit critical damage indicators (Qwen2-VL). 

In Scenario (d), ground truth demonstrates superior completeness by explicitly mentioning "patches of bare earth," "fewer structures," and reduced boats, which align with visible changes in the images (e.g., exposed soil, collapsed buildings). While both captions lack explicit disaster type identification, ground truth’s specificity on environmental and structural impacts ("significant clearing," "shift in human activity") enhances accuracy and clarity . LLaVA-Interleave’s vague reference to "debris" and omission of key details (e.g., bare earth) makes it less precise. Both adhere to facts, but ground truth is richer detail elevates its overall quality.

Figure~\ref{fig:more_examples} shows more samples on RSCC subset along with baseline results. Table~\ref{tab:captioning_results_all} shows overall quantitative results on RSCC subset. It is witnessed that auxiliary building damage info augmentation greatly improve the quality of change captions. We also find performance gets saturated equipped with auxiliary info regardless model size. We provide an additional metric BLEURT~\footnote{\url{https://huggingface.co/lucadiliello/BLEURT-20-D12}}~\citep{Sellam_Das_Parikh_2020}, a learned evaluation metric to measure contextual similarity as well. However, the BLEURT is strongly biased on text length, which fails in valid evaluation. We are seeking for more reliable metrics in the future. Table~\ref{tab:disaster_events} and ~\ref{tab:model_config} display RSCC data source details and baseline model configurations respectively.

\begin{table}[!h]
\caption{Detailed image caption performance on the subset of RSCC dataset. Avg\_L denotes the average word number of generated captions. \textbf{Boldface} indicates the best performance while \underline{underline} denotes the suboptimal performance.$^*$We observe that BLIP-3 (XGen-MM) and LLaVA-OneVision tend to repeat their answer endlessly, which cause large caption lengths.} 
\vspace{0.5em}
\centering
\small
\setlength{\tabcolsep}{0.7pt}
\renewcommand{\arraystretch}{1.2}
\begin{tabular}{lllllr}
\toprule
\multicolumn{1}{c}{\textbf{Model}} & 
\multicolumn{2}{c}{\textbf{N-Gram}} & 
\multicolumn{2}{c}{\textbf{Contextual Similarity}} & 
\multirow{2}{*}{\textbf{Avg\_L}} \\
\cmidrule(lr){2-3} \cmidrule(lr){4-5}
\multicolumn{1}{c}{\textbf{(\#activate params)}} & 
\textbf{ROUGE(\%)↑} & 
\textbf{METEOR(\%)↑} &
\textbf{BLEURT(\%)↑} &
\textbf{ST5-SCS(\%)↑} & 
 \\
\midrule
BLIP-3 (3B)~\citep{xueXGenMMBLIP3Family2024} & 4.53 & 10.85 & 56.49 & 44.05 & $^*$456 \\
{\color{tech_purple}\hspace{1.5mm}+ Textual Prompt} & 10.07 (\textcolor{green}{+5.54↑}) & 20.69 (\textcolor{green}{+9.84↑}) & 56.79 (\textcolor{green}{+0.30↑}) & 63.67 (\textcolor{green}{+19.62↑}) & $^*$302 \\
{\color{tech_purple}\hspace{5mm}+ Visual Prompt} & 8.45 (\textcolor{red}{-1.62↓}) & 19.18 (\textcolor{red}{-1.51↓}) & 60.24 (\textcolor{green}{+3.45↑}) & 68.34 (\textcolor{green}{+4.67↑}) & $^*$354 \\
Kimi-VL (3B)~\citep{kimiteam2025kimivltechnicalreport} & 12.47 & 16.95 & 45.11 & 51.35 & 87 \\
{\color{tech_purple}\hspace{1.5mm}+ Textual Prompt} & 16.83 (\textcolor{green}{+4.36↑}) & 25.47 (\textcolor{green}{+8.52↑}) & 54.55 (\textcolor{green}{+9.44↑}) & 70.75 (\textcolor{green}{+19.40↑}) & 108 \\
{\color{tech_purple}\hspace{5mm}+ Visual Prompt} & 16.83 (\textcolor{black}{+0.00}) & 25.39 (\textcolor{red}{-0.08↓}) & 54.24 (\textcolor{red}{-0.31↓}) & 69.97 (\textcolor{red}{-0.78↓}) & 109 \\
Phi-4-Multimodal (4B)~\citep{microsoftPhi4MiniTechnicalReport2025} & 4.09 & 1.45 & 23.51 & 34.55 & 7 \\
{\color{tech_purple}\hspace{1.5mm}+ Textual Prompt} & 17.08 (\textcolor{green}{+13.00↑}) & 19.70 (\textcolor{green}{+18.25↑}) & 52.00 (\textcolor{green}{+28.49↑}) & 67.62 (\textcolor{green}{+33.07↑}) & 75 \\
{\color{tech_purple}\hspace{5mm}+ Visual Prompt} & 17.05 (\textcolor{red}{-0.03↓}) & 19.09 (\textcolor{red}{-0.61↓}) & 51.46 (\textcolor{red}{-0.54↓}) & 66.69 (\textcolor{red}{-0.93↓}) & 70 \\
Qwen2-VL (7B)~\citep{wangQwen2VLEnhancingVisionLanguage2024} & 11.02 & 9.95 & 38.86 & 45.55 & 42 \\
{\color{tech_purple}\hspace{1.5mm}+ Textual Prompt} & 19.04 (\textcolor{green}{+8.02↑}) & 25.20 (\textcolor{green}{+15.25↑}) & 52.64 (\textcolor{green}{+13.78↑}) & 72.65 (\textcolor{green}{+27.10↑}) & 84 \\
{\color{tech_purple}\hspace{5mm}+ Visual Prompt} & 18.43 (\textcolor{red}{-0.61↓}) & 25.03 (\textcolor{red}{-0.17↓}) & 52.27 (\textcolor{red}{-0.37↓}) & 72.89 (\textcolor{green}{+0.24↑}) & 88 \\
LLaVA-NeXT-Interleave (8B)~\citep{liLLaVANeXTInterleaveTacklingMultiimage2024} & 12.51 & 13.29 & 42.80 & 46.99 & 57 \\
{\color{tech_purple}\hspace{1.5mm}+ Textual Prompt} & 16.09 (\textcolor{green}{+3.58↑}) & 20.73 (\textcolor{green}{+7.44↑}) & 50.01 (\textcolor{green}{+7.21↑}) & 62.60 (\textcolor{green}{+15.61↑}) & 75 \\
{\color{tech_purple}\hspace{5mm}+ Visual Prompt} & 15.76 (\textcolor{red}{-0.33↓}) & 21.17 (\textcolor{green}{+0.44↑}) & 50.08 (\textcolor{green}{+0.07↑}) & 65.75 (\textcolor{green}{+3.15↑}) & 88 \\
LLaVA-OneVision (8B)~\citep{liLLaVAOneVisionEasyVisual2024a} & 8.40 & 10.97 & 46.27 & 46.15 & $^*$221 \\
{\color{tech_purple}\hspace{1.5mm}+ Textual Prompt} & 11.15 (\textcolor{green}{+2.75↑}) & 19.09 (\textcolor{green}{+8.12↑}) & \textbf{61.37} (\textcolor{green}{+15.10↑}) & 70.08 (\textcolor{green}{+23.93↑}) & $^*$285 \\
{\color{tech_purple}\hspace{5mm}+ Visual Prompt} & 10.68 (\textcolor{red}{-0.47↓}) & 18.27 (\textcolor{red}{-0.82↓}) & \underline{60.59} (\textcolor{red}{-0.78↓}) & 69.34 (\textcolor{red}{-0.74↓}) & $^*$290 \\
InternVL 3 (8B)~\citep{Internvl3} & 12.76 & 15.77 & 43.97 & 51.84 & 64 \\
{\color{tech_purple}\hspace{1.5mm}+ Textual Prompt} & \underline{19.81} (\textcolor{green}{+7.05↑}) & \underline{28.51} (\textcolor{green}{+12.74↑}) & 56.51 (\textcolor{green}{+12.54↑}) & 78.57 (\textcolor{green}{+26.73↑}) & 81 \\
{\color{tech_purple}\hspace{5mm}+ Visual Prompt} & 19.70 (\textcolor{red}{-0.11↓}) & 28.46 (\textcolor{red}{-0.05↓}) & 56.10 (\textcolor{red}{-0.41↓}) & \textbf{79.18} (\textcolor{green}{+0.61↑}) & 84 \\
Pixtral (12B)~\citep{agrawalPixtral12B2024} & 12.34 & 15.94 & 43.74 & 49.36 & 70 \\
{\color{tech_purple}\hspace{1.5mm}+ Textual Prompt} & \textbf{19.87} (\textcolor{green}{+7.53↑}) & \textbf{29.01} (\textcolor{green}{+13.07↑}) & 55.79 (\textcolor{green}{+12.05↑}) & \underline{79.07} (\textcolor{green}{+29.71↑}) & 97 \\
{\color{tech_purple}\hspace{5mm}+ Visual Prompt} & 19.03 (\textcolor{red}{-0.84↓}) & 28.44 (\textcolor{red}{-0.57↓}) & 54.99 (\textcolor{red}{-0.80↓}) & 78.71 (\textcolor{red}{-0.36↓}) & 102 \\
\textcolor{gray}{CCExpert (7B)}~\citep{wangCCExpertAdvancingMLLM2024} & 7.61 & 4.32 & 35.21 & 40.81 & 12 \\
{\color{tech_purple}\hspace{1.5mm}{+ Textual Prompt}} & 8.71 (\textcolor{green}{+1.10↑}) & 5.35 (\textcolor{green}{+1.03↑}) & 39.01 (\textcolor{green}{+3.80↑}) & 47.13 (\textcolor{green}{+6.32↑}) & 14 \\
{\color{tech_purple}\hspace{5mm}{+ Visual Prompt}} & 8.84 (\textcolor{green}{+0.13↑}) & 5.41 (\textcolor{green}{+0.06↑}) & 38.94 (\textcolor{red}{-0.07↓}) & 46.58 (\textcolor{red}{-0.55↓}) & 14 \\
\textcolor{gray}{TEOChat (7B)}~\citep{irvinTEOChatLargeVisionLanguage2024a} & 7.86 & 5.77 & 39.47 & 52.64 & 15 \\
{\color{tech_purple}\hspace{1.5mm}{+ Textual Prompt}} & 11.81 (\textcolor{green}{+3.95↑}) & 10.24 (\textcolor{green}{+4.47↑}) & 45.53 (\textcolor{green}{+6.06↑}) & 61.73 (\textcolor{green}{+9.09↑}) & 22 \\
{\color{tech_purple}\hspace{5mm}{+ Visual Prompt}} & 11.55 (\textcolor{red}{-0.26↓}) & 10.04 (\textcolor{red}{-0.20↓}) & 45.31 (\textcolor{red}{-0.22↓}) & 62.53 (\textcolor{green}{+0.80↑}) & 22 \\
\textcolor{gray}{Ours (7B)} & 14.99 & 16.05 & 45.50 & 58.52 & 44 \\
{\color{tech_purple}\hspace{1.5mm}{+ Textual Prompt}} & 22.23 (\textcolor{green}{+7.24↑}) & 33.83 (\textcolor{green}{+17.78↑}) & 56.87 (\textcolor{green}{+11.37↑}) & 78.02 (\textcolor{green}{+19.50↑}) & 76 \\
{\color{tech_purple}\hspace{5mm}{+ Visual Prompt}} & 22.37 (\textcolor{green}{+0.14↑}) & 33.81 (\textcolor{red}{-0.02↓}) & 57.02 (\textcolor{green}{+0.15↑}) & 78.87 (\textcolor{green}{+0.85↑}) & 79 \\
Qwen2.5-VL (72B)~\citep{baiQwen25VLTechnicalReport2025} & - & - & - & - & - \\
{\color{tech_purple}\hspace{1.5mm}+ Textual Prompt} & - & - & - & 76.84 & 53 \\
{\color{tech_purple}\hspace{5mm}+ Visual Prompt} & - & - & - & 76.85 & 57 \\

\bottomrule
\end{tabular}
\label{tab:captioning_results_all}
\end{table}

\begin{figure}[!ht]
    \centering
    \includegraphics[width=0.7\linewidth]{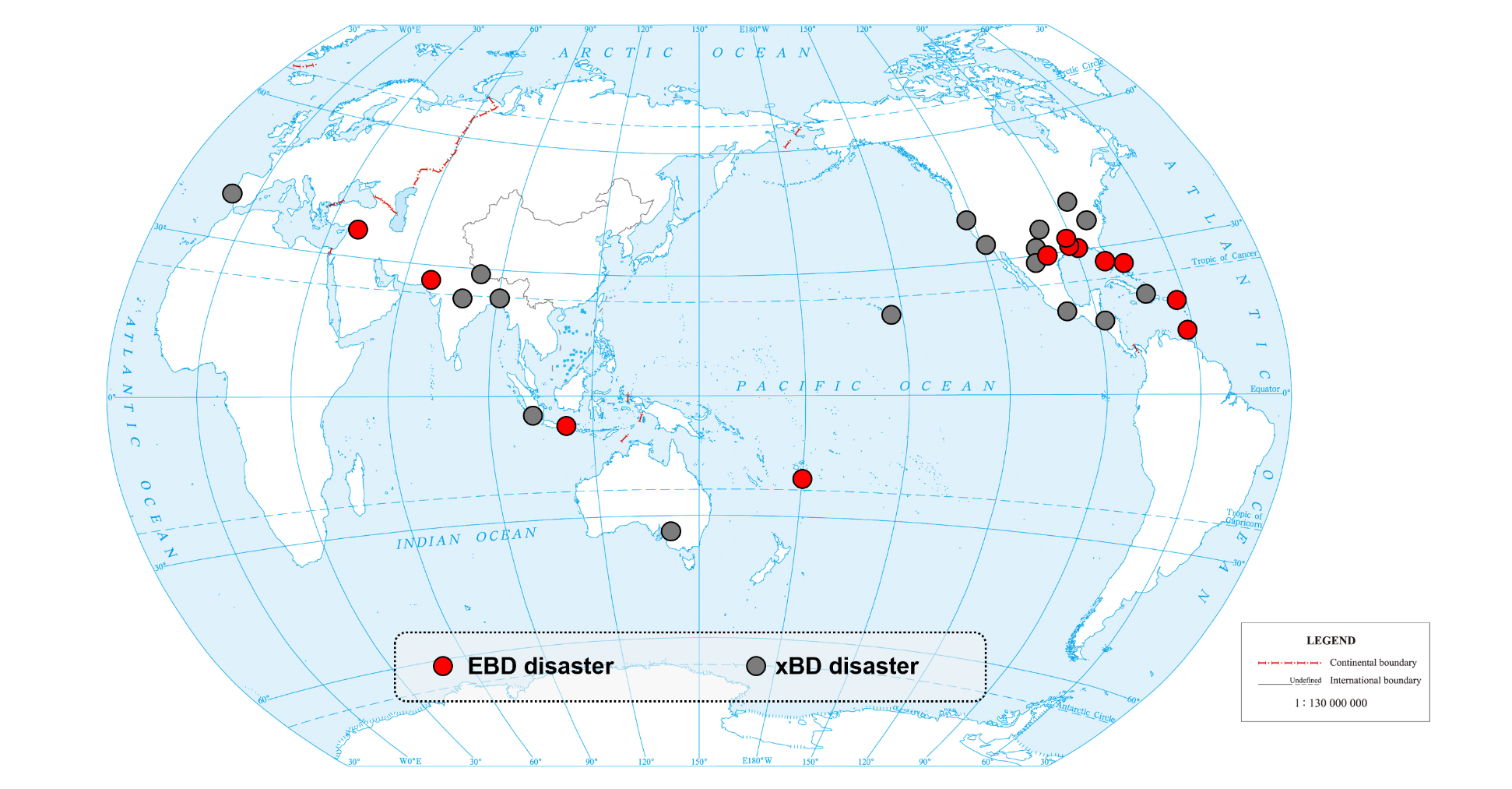}
    \caption{RSCC (EBD + xBD) distribution. Image Credit: Wang~\textit{et  al.}~\citep{Wang2024EBD}.}
    \label{fig:rscc_distribution}
\end{figure}

\begin{table}[!b]
\caption{The 31 disaster events from RSCC dataset.}
\vspace{0.5em}
\centering
\footnotesize
\renewcommand{\arraystretch}{1.3}
\label{tab:disaster_events}
\begin{tabular}{c|lll}
\toprule
\textbf{Source} & \textbf{Disaster type} & \textbf{Disaster event} & \textbf{Event date} \\
\midrule
\multirow{19}{*}{\textbf{xBD}} 
& Earthquake & Mexico City earthquake & Sep 19, 2017 \\
& Wildfire & Portugal wildfires & Jun 17-24, 2017 \\
& Wildfire & Santa Rosa wildfires & Oct 8-31, 2017 \\
& Wildfire & Carr wildfire & Jul 23-Aug 30, 2018 \\
& Wildfire & Woolsey fire & Nov 9-28, 2018 \\
& Wildfire & Pinery fire & Nov 25-Dec 2, 2018 \\
& Volcano & Lower Puna volcanic eruption & May 23-Aug 14, 2018 \\
& Volcano & Guatemala Fuego volcanic eruption & Jun 3, 2018 \\
& Storm & Tuscaloosa, AL tornado & Apr 27, 2011 \\
& Storm & Joplin, MO tornado & May 22, 2011 \\
& Storm & Moore, OK tornado & May 20, 2013 \\
& Storm & Hurricane Matthew & Sep 28-Oct 10, 2016 \\
& Storm & Hurricane Florence & Sep 10-19, 2018 \\
& Flooding & Monsoon in Nepal, India, Bangladesh & Jul-Sep, 2017 \\
& Flooding & Hurricane Harvey & Aug 17-Sep 2, 2017 \\
& Flooding & Hurricane Michael & Oct 7-16, 2018 \\
& Flooding & Midwest US floods & Jan 3-May 31, 2019 \\
& Tsunami & Indonesia tsunami & Sep 18, 2018 \\
& Tsunami & Sunda Strait tsunami & Dec 22, 2018 \\
\cline{2-4}
\multirow{12}{*}{\textbf{EBD}}
& Hurricane & Hurricane Delta & Oct 8, 2020 \\
& Hurricane & Hurricane Dorian & Sep 1, 2019 \\
& Hurricane & Hurricane Ida & Oct 29, 2021 \\
& Hurricane & Hurricane Laura & Aug 26, 2020 \\
& Hurricane & Hurricane Irma & Sep 6, 2017 \\
& Hurricane & Hurricane Ian & Sep 26, 2022 \\
& Tornadoes & Texas Tornadoes & Mar 23, 2022 \\
& Volcanic Eruption & Mount Semeru Eruption & Dec 4, 2021 \\
& Volcanic Eruption & ST. Vincent Volcano & Apr 9, 2021 \\
& Volcanic Eruption & Tonga Volcano & Jan 15, 2022 \\
& Earthquake & Turkey Earthquake & Feb 6, 2023 \\
& Flooding & Pakistan Flooding & Jul 26, 2022 \\
\bottomrule
\end{tabular}
\end{table}

\begin{table}[!ht]
\renewcommand{\arraystretch}{1.3}
\caption{Configuration of baseline models.}
\vspace{0.5em}
\centering
\label{tab:model_config}
\begin{tabular}{lcll}
\hline
\textbf{Model Name} & \textbf{\#Active Parameters} & \textbf{LLM} & \textbf{Image Encoder} \\
\hline
Kimi-VL & 3B & Moonlight-A3B-E18B & MoonViT \\
BLIP-3 & 4B & Phi-3-mini-4B & SigLIP \\
Phi-4-Multimodal & 4B & Phi-4-Mini 4B & SigLIP (LORA) \\
LLaVA-NeXT-Interleave & 7B & Qwen1.5 7B & SigLIP \\
Qwen2-VL & 7B & Qwen2-7B & DFN's ViT-H \\
LLaVa-OneVision & 7B & Qwen2 7B & SigLIP \\
InternVL 3 & 8B & Qwen2.5-7B & InternViT-300M \\
Pixtral & 12B & Mistral-Nemo-12B & PixtralViT \\
\hline
TEOChat & 7B & Vicuna-v1.5-7B & OpenCLIP-L/14 \\
CCExpert & 7B & Qwen2-7B & SigLIP \\
\hline
\end{tabular}
\end{table}

\clearpage
\subsection{Access to Data}

The RSCC dataset can be accessed and downloaded through our dedicated platform, which provides detailed views of the dataset components and their annotations.  For practical examples and to download the dataset, visit our Huggingface repository (\url{https://huggingface.co/BiliSakura/RSCC}). Detailed metadata for the dataset is documented using the Croissant metadata framework, ensuring comprehensive coverage and compliance with the MLCommons Croissant standards, check [metadata](\url{https://huggingface.co/api/datasets/BiliSakura/RSCC}). Please check our Huggingface repo for metadata details. We also release our specialized model RSCCM (\url{https://huggingface.co/api/models/BiliSakura/RSCCM}).

\subsection{Author Statement and Data License}
\textbf{Author Responsibility Statement:} The authors bear all responsibilities in case of any violations of rights or ethical concerns regarding the RSCC dataset.
\newline
\textbf{Data License Confirmation:} The dataset is released under the [CC-BY-4.0], which permits unrestricted use, distribution, and reproduction in any medium, provided the original work is properly cited.

\subsection{Broader Impacts}
The dataset consists of non-sensitive, publicly available satellite images where no individual person or private property can be identified. Users are encouraged to use RSCC responsibly and ethically, particularly when developing applications that might impact environmental monitoring and urban planning.

\subsection{Prompt Template}
\label{template}

\begin{tcolorbox}[
    colback=black!5!white,
    colframe=black!75!black,
    title=Prompt Template (Post Correction based on metadata)
]
You will be provided with a change caption of a pair of remote sensing images, and metadata containing building damage statistics and disaster type. Perform the following analysis:

\begin{enumerate}
    \item \textbf{Disaster Type Inference:} 
    
- Determine the disaster type (e.g., flood, wildfire) based on textual context.

    \item \textbf{Keyword Evaluation:}

- Extract disaster-relevant keywords from the caption.
        
- Ensure these keywords are logically consistent with the inferred disaster type.

    \item \textbf{Damage Statistics Validation:}

- \textbf{Counts:} Compare the number of buildings per damage level in the caption (e.g., ''24 minor-damaged'') with the metadata values (e.g., \texttt{\{"minor-damage": 24\}}).
        
- \textbf{Levels:} Verify that damage level terms (e.g., ''destroyed'' vs. ''major-damage'') match the metadata's labeling scheme.

    \item \textbf{Flag Mismatches:}

- \textbf{Keyword Mismatch:} Keywords incompatible with the disaster type (e.g., ''volcanic ash'' in a flood caption).
        
- \textbf{Count Mismatch:} Discrepancies between caption and metadata (e.g., ''24 minor-damaged'' vs. metadata \texttt{\{"minor-damage": 20\}}).
        
- \textbf{Level Mismatch:} Incorrect damage level terminology (e.g., ''severe'' instead of ''major-damage'').

    \item \textbf{Return:}

- ''PASS'' if all criteria are met.
        
- ''FAIL'' with specific violations (e.g., ''Count mismatch: Minor-damaged (caption:24 vs. metadata:20); Level mismatch: 'severe' instead of 'major-damage'\ '').
    
\end{enumerate}
\end{tcolorbox}

\clearpage
\begin{tcolorbox}[
    colback=white, 
    colframe=black!75!black, 
    title=Prompt Templates,
    subtitle style={colback=black!50!white, colframe=black!50!black} %
]
\tcbsubtitle{Prompt Template (Zero-Shot)}
<image>\textbackslash n<image>\textbackslash n
Give change description between two satellite images. Output answer in a news style with a few sentences using precise phrases separated by commas.

\tcbsubtitle{Prompt Template (Textual Prompt)}
<image>\textbackslash n<image>\textbackslash nThese two satellite images show a \textcolor{cyan}{\{disaster\_type\}} natural disaster. Here is the disaster level descriptions:
\begin{itemize}
  \item Disaster Level 0 (No Damage): Undisturbed. No sign of water, structural or shingle damage, or burn marks. 
  \item Disaster Level 1 (Minor Damage): Building partially burnt, water surrounding structure, volcanic flow nearby, roof elements missing, or visible cracks. 
  \item Disaster Level 2 (Major Damage): Partial wall or roof collapse, encroaching volcanic flow, or surrounded by water/mud. 
  \item Disaster Level 3 (Destroyed): Scorched, completely collapsed, partially/completely covered with water/mud, or otherwise no longer present. 
\end{itemize}
 We already know that there are \textcolor{cyan}{\{number[all]\}} buildings. \textcolor{cyan}{\{number[no-damage]\}} buildings are no damaged. \textcolor{cyan}{\{number[minor-damage]\}} buildings are minor damaged,  \textcolor{cyan}{\{number[major-damage]\}} building are major damaged, \textcolor{cyan}{\{number[destroyed]\}} buildings are destroyed. \textcolor{cyan}{\{number[unclassified]\}} buildings damage are unknown due to some reasons. Now, describe the changes that occurred between the pre-event and post-event images in a news style with the given disaster level descriptions.

\tcbsubtitle{Prompt Template (Visual Prompt)}
<image>\textbackslash n<image>\textbackslash nThese two satellite images show a \textcolor{cyan}{\{disaster\_type\}} natural disaster. Here is the disaster level descriptions:
\begin{itemize}
  \item Disaster Level 0 (No Damage): Undisturbed. No sign of water, structural or shingle damage, or burn marks. 
  \item Disaster Level 1 (Minor Damage): Building partially burnt, water surrounding structure, volcanic flow nearby, roof elements missing, or visible cracks. 
  \item Disaster Level 2 (Major Damage): Partial wall or roof collapse, encroaching volcanic flow, or surrounded by water/mud. 
  \item Disaster Level 3 (Destroyed): Scorched, completely collapsed, partially/completely covered with water/mud, or otherwise no longer present. 
\end{itemize}
 We already know that there are \textcolor{cyan}{\{number[all]\}} buildings. \textcolor{cyan}{\{number[no-damage]\}} buildings are no damaged colored in \textcolor{green}{green}. \textcolor{cyan}{\{number[minor-damage]\}} buildings are minor damaged colored in \textcolor{blue}{blue},  \textcolor{cyan}{\{number[major-damage]\}} building are major damaged colored in \textcolor{orange}{orange}, \textcolor{cyan}{\{number[destroyed]\}} buildings are destroyed colored in \textcolor{red}{red}. \textcolor{cyan}{\{number[unclassified]\}} buildings damage are unknown due to some reasons colored in \textcolor{white}{white}. Now, describe the changes that occurred between the pre-event and post-event images in a news style with the given disaster level descriptions.
\end{tcolorbox}

\subsection{Details of Human Preference Study}

\begin{tcolorbox}[colback=black!5!white,colframe=black!75!black,title=Human Preference Guidelines]
You will be provided with 2 satellite images of the same area before and after a natural disaster event.  Your task is to evaluate change captions generated by different vision language models and select the best one.

Evaluation Criteria:
\begin{enumerate}
    \item Accuracy - Correct interpretation of damage patterns and disaster type
    \item Completeness - Inclusion of relevant details (structures affected, disaster indicators)
    \item Clarity - Clear, concise description without contradictions
    \item Adherence to Facts - Consistency with typical disaster damage level
\end{enumerate}

Follow the criteria and choose the best change caption by click the corresponding radio button.

\end{tcolorbox}

\begin{figure*}[!ht]
\centering
\includegraphics[width=1.0\linewidth]{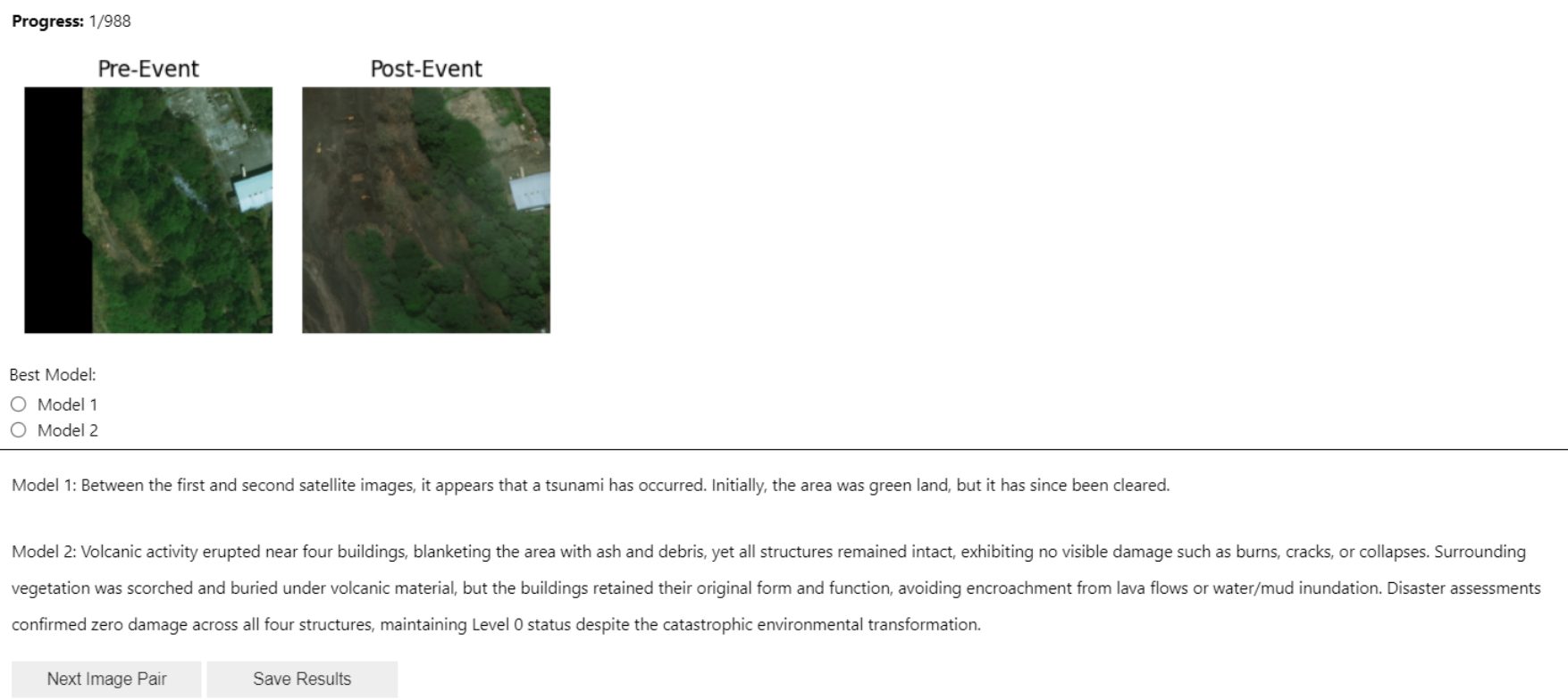}
\caption{A screenshot of human preference study labeling interface.}
\label{fig:screenshot}
\end{figure*} 


\end{document}